\documentclass[a4paper,11pt]{article}
\usepackage{url}
\usepackage{graphicx}
\usepackage{natbib}
\bibpunct{(}{)}{,}{a}{,}{,}
\usepackage{amsmath}
\usepackage[labelfont=bf]{caption}
\usepackage[labelfont=bf]{subcaption}
\usepackage{amsfonts}
\usepackage{commath}
\usepackage{ctable}
\usepackage{color,colortbl,array}
\definecolor{Gray}{gray}{0.9}
\usepackage{afterpage}
\usepackage{subcaption}
\usepackage[a4paper,margin=1in]{geometry}
\usepackage{tabularx}
\usepackage{hyperref}

\author{Fabi\'an Slonimczyk\thanks{International College of Economics and Finance, HSE. \texttt{fslonimczyk@hse.ru}.}}
\title{This Candidate is [MASK]\\Prompt-based Sentiment Extraction and Reference Letters}
\date{October 2025}

\begin{document}
\bibliographystyle{kbib}

\maketitle

\begin{abstract}
I propose a relatively simple way to deploy pre-trained large language models (LLMs) in order to extract sentiment and other useful features from text data. The method, which I refer to as \emph{prompt-based sentiment extraction}, offers multiple advantages over other methods used in economics and finance. In particular, it accepts the text input as is (without pre-processing) and produces a sentiment score that has a probability interpretation. Unlike other LLM-based approaches, it does not require any fine-tuning or labeled data. I apply my prompt-based strategy to a hand-collected corpus of confidential reference letters (RLs). I show that the sentiment contents of RLs are clearly reflected in job market outcomes. Candidates with higher average sentiment in their RLs perform markedly better regardless of the measure of success chosen. Moreover, I show that sentiment dispersion among letter writers negatively affects the job market candidate's performance. I compare my sentiment extraction approach to other commonly used methods for sentiment analysis: `bag-of-words' approaches, fine-tuned language models, and querying advanced chatbots. No other method can fully reproduce the results obtained by prompt-based sentiment extraction. Finally, I slightly modify the method to obtain `gendered' sentiment scores \citep[as in][]{Eberhardt_et_al_2023}. I show that RLs written for female candidates emphasize `grindstone' personality traits, whereas male candidates' letters emphasize `standout' traits. These gender differences negatively affect women's job market outcomes.
\end{abstract}

\begin{flushleft}
{\footnotesize JEL classification: C45; J16\newline
 Key words: Large language models; text data; sentiment analysis; reference letters. }
\end{flushleft}

\setlength{\parskip}{0.1cm}
\newcommand{\cov}{\mathrm{cov}}
\newcommand{\1}{\mathbf{1}}
\newcommand{\E}{\mathbf{E}}
\newcommand{\D}{\mathbf{\Delta}}

\newpage
\section{Introduction}

The idea that sentiment---subjective states of mind and opinions---can be an important determinant of economic decisions and outcomes is widely accepted both in micro and macroeconomic settings \citep[e.g.][]{IzmalkovYildiz_2010,Angeletos_et_al_2013,Angeletos_et_al_2018,Maxted_2024}. It is not surprising, therefore, that considerable effort has been exerted by researchers trying to measure it. There are myriad applications of sentiment analysis in economics: in the context of business cycles \citep{Shapiro_et_al_2022}, policy-level uncertainty \citep{Baker_et_al_2016,Caldara_Iacoviello_2022}, the effects of monetary policy \citep{TerEllen_et_al_2022}, and asset pricing \citep{Garcia_2013}, just to name a few.

In this paper, I propose a relatively simple way to deploy pre-trained large language models (LLMs) in order to extract sentiment and other useful features from text data. I refer to this method as \emph{prompt-based sentiment extraction}. In a nutshell, the method involves combining the input text with a suitable prompt, so that the LLM's emergent abilities acquired during pre-training can be applied to the sentiment extraction task. For example, in order to recognize the emotion of the social media post ``I missed the bus today!'' we may append the prompt ``I felt so [MASK]'' and ask the LLM to fill the blank (denoted by the [MASK] token) with an emotion-bearing word. The LLM's output, consisting of a probability distribution over its vocabulary, can then be summarized to produce the desired sentiment score.

This relatively simple method has multiple advantages over alternative techniques. Until fairly recently, the leading approaches to sentiment extraction in economics were different versions of the ``bag-of-words'' method, in which the order of the terms in the text do not affect the resulting sentiment score (only the presence or absence of certain words matters). In sharp contrast, within LLMs words and other language parts receive a representation that is highly dependent on the context in which the term is used. As a simple example, LLMs can distinguish without difficulty between uses of the word ``fly'' that refer to an insect from cases when it is used as a verb. The highly complex sequential relationships among words, clauses, sentences and paragraphs is accepted without arbitrary restrictions and in turn used to infer the correct representation of each term in the sequence.

Secondly, unlike past methods that required heavy-handed preprocessing of the text input, LLMs accept natural language ``as is'', inclusive of punctuation signs and other special characters. The vocabulary size of recent LLMs is truly staggering. For example, the main model I use in my application is \texttt{Llama 3.1}, which has a vocabulary size of over 120 thousand tokens. For comparison, Shakespeare used around 20 thousand words in his plays and poems. In addition, Llama and other state-of-the-art LLMs are multi-lingual, accepting words in Spanish, Portuguese, Italian, German, Thai, French, and Hindi aside from English \citep{Llamablog_2024}.

A third key advantage of prompt-based sentiment extraction is that the resulting sentiment scores have a probability interpretation. Specifically, the polarity sentiment score that I define is simply the difference between the probabilities that the LLM assigns to positive and negative completions of the prompt. This makes the interpretation of downstream applications of the scores---e.g. in regression analysis---much easier. Fourth, the LLM's ability to correctly interpret the text can be directly assessed by testing its predictions against known text meta-data. For example, I show below that the LLM performs extremely well in predicting the job market candidate's sex and field of research by simply ``reading'' the RLs. While this is not strict proof of the accuracy of the sentiment scores, it is arguably superior to the use of completely untestable alternatives. More generally, prompt-based learning can be used to extract features of text other than sentiment. For example, in this paper I use the technique to extract measures of how much the recommendation letter writers emphasize `grindstone' and `standout' aspects of the candidate's personality.

I apply prompt-based sentiment extraction to a hand-collected corpus of confidential RLs submitted as a standard part of the package that academic economists send to potential employers in academia and other sectors. Letters of recommendation have a number of desirable features that make them an ideal setting to test my sentiment extraction method. Because they are confidential, we can be sure that they have not been used as part of the training set for any of the open-source LLMs that I test-drive. As argued recently by \citet{Ludwig_et_al_2025}, the existence of \emph{leakage} between the LLM's training data and the economic application has the potential to severely bias results. In addition, RLs provide an interesting mix of relatively objective/factual descriptions and more opinionated content. While the structure of the letters is relatively stable, the opinions and discussion can often be quite idiosyncratic, so the task of producing a sentiment score is far from trivial. Finally, on average letters are around 2,000 words long, which is relatively easy to process for state-of-the-art LLMs.

Besides showcasing prompt-based sentiment extraction, the application is meant to contribute to the growing literature on the role of references in the labor market, and how apparently innocuous small biases may contribute to gender inequality. The economics job market is an ideal setting to understand how references are used as a way to screen applicants. Because all job market candidates have similar formal credentials (a PhD), the use of education as a signalling device is implicitly controlled for. As a way to further control for the level of training, I focus only on first-time job market participants. In other words, I only look at job market candidates who are close to finishing their PhD program and have not held a permanent academic position before.

The job market for economists is somewhat unique in that all open positions are widely advertised. With few exceptions, all openings are posted in a central database accessible to all market participants. The application process itself has become highly standardized. Candidates are expected to submit a CV, a research sample (``job market paper''), and at least two confidential letters of reference.\footnote{Some employers may require additional materials. For example, some colleges with teaching-focused positions often require a ``teaching philosophy'' statement and the submission of student evaluations.} Because the letter writers typically submit the same letters to all positions the job market candidates apply to \citep{Coles_et_al_2010}, it is fair to assume that I have precise information on who wrote letters for each job market candidate in my sample and what the contents of those letters are.\footnote{Having precise information on both the referee and the referred person is very rare. Most prior studies of referrals in the labor market do not observe recommendations directly and are therefore forced to use some kind of proxy. For example, \citet{Dustmann_et_al_2016} define referral networks in terms of ethnicity, while \citet{Bayer_et_al_2008} and \citet{Hellerstein_et_al_2011} use the neighborhood of residence as proxy.} Finally, the extensive use of electronic applications means that applying for a job opening is almost costless to the candidate. In summary, the specific characteristics of the market mean that, unlike other settings, here RLs are \textbf{not} being used to disseminate information about job market opportunities.\footnote{For a model of the role of social networks as a manner of obtaining information about job opportunities, see \citet{Calvo-Armengol_Jackson_2004}.} Instead, reference letters are primarily being used as a way to screen applicants.

The issue of under what conditions indirect signaling via letters of recommendation can function efficiently has been a matter of concern in the theoretical literature. Because referees compete against each other, we can expect them to make excessive claims about the candidates they are recommending in order to have them placed ahead of the candidates being recommended by rival referees \citep{Saloner_1985}.\footnote{Other models in which referrals serve as a screening mechanism to reduce the information asymmetry inherent to the hiring process are \citet{Montgomery_1991} and \citet{Simon_Warner_1992}.} Moreover, referees' motivation for referring applicants and providing truthful recommendations is not well understood. \citet{Camara_et_al_2023} model the reputation game facing referees explicitly. Remarkably, they find that referees may actively lie about the candidate's true ability in equilibrium, and that the extent of the lying depends on the intricate interplay between preferences, reputation and lying costs.\footnote{Another paper that investigates referees' incentives theoretically is \citet{Ekinci_2016}. \citet{Carmichael_1988} argues that the reason why universities have instituted a tenure system in academic departments is precisely to ensure that faculty members are properly incentivized to hire the best junior job market candidates.}

In line with these considerations, the extent to which the strength of an applicant's references correlate with good labor market outcomes is an empirical question without an obvious a priori answer. If hiring committees learn to distrust the contents of reference letters because of generalized ``recommendation inflation,'' then the correlation might be statistically indistinguishable from zero. On the other hand, if reputational and other costs are severe enough to align the incentives of referees with those of hiring committees, then we can expect to find that good references more often than not translate into success in the job market.

There are several empirical studies of how references affect labor market outcomes. Using a long panel data set covering all hiring activity of a single US corporation, \citet{Brown_et_al_2016} found that referred candidates are more likely to be hired than non-referred candidates. \citet{Burks_et_al_2015} assembled personnel data from nine large firms in the call center, trucking, and high-tech industries. Consistent with \citet{Brown_et_al_2016}, they found that referred applicants are substantially more likely to be hired, and that, conditional on receiving an offer, they are more likely to accept it.

There is also some experimental evidence on the value of reference letters. \citet{Pallais_2014} conducted a field experiment on an online marketplace in which workers bid to be hired for short term employment. One of her treatment groups received detailed feedback after completing a data-entry task, while another group received only a one-to-five rating. She found that both treatment groups benefitted from treatment relative to a control group that was not hired. \citet{Abel_et_al_2020} conducted an audit study where they randomly appended a reference letter to some applications submitted on behalf of job seekers to vacancies in South Africa. They found that, for the same applicant, attaching a letter increased the probability of receiving a response and the rate of interview requests by about 60 percent.

A separate empirical literature has investigated whether there are differences in the language chosen to describe female and male candidates. Starting with the work of \citet{Trix_Psenka_2003} and \citet{Schmader_et_al_2007}, a number of papers have documented that women are described with more `grindstone' attributes and fewer `ability' or `standout' ones. Two papers study the use of reference letters in the economics job market \citep{Eberhardt_et_al_2023,Baltrunaite_et_al_2024}. Their data is similar to mine in that they collect administrative hiring data from institutions hiring in the international economics job market.\footnote{\citet{Eberhardt_et_al_2023} study a sample of applicants to a top UK economics department, while \citet{Baltrunaite_et_al_2024} look at a premier economics department and a top policy institution in Italy.} They both document similar differences in language across gender lines using the lexical approach. In this paper, I confirm their findings using a prompt-based strategy.

In summary, I find that prompt-based sentiment extraction produces sentiment scores that are strongly correlated with the probability of obtaining a good job market outcome. In particular, high average sentiment among all referees are predictive of a sizeable higher probability of success. Moreover, my paper is the first to document that disagreement among letter writers can offset the positive effect of high average RL sentiment, which is strictly consistent with employers taking all available information in RLs pretty much at face value. While these findings are not causal, they strongly suggest that the theoretical literature might have over-emphasized the potential pitfalls of RLs.

Until fairly recently, feasible strategies to take advantage of text data were rather limited. In the next section, I provide a brief review of the existing methods and a formal description of my prompt-based strategy. I then move to a description of the data, including a descriptive analysis of prompt-based sentiment scores obtained from RLs. Section~4 presents the main results linking RL sentiment and job market outcomes. In section~5, I present results related to gender differences in the language used by letter writers to describe candidates. Section~6 concludes.

\section{Sentiment Analysis and Prompt-based Learning}
\label{section_methods}

In this paper, I present a prompt-based methodology to extract sentiment and other scores from text data. This approach has several advantages over other methods. In this section, I first provide a brief description of the lexical approach, which has been the most commonly applied method to assign a sentiment score to text in the economics and finance literature. I then briefly describe how the development of language models has revolutionized sentiment analysis and formally present the prompt-based methodology for sentiment extraction.\footnote{My description of other methods is limited to the key elements necessary for a comparison with prompt-based sentiment extraction. The classic \citet{Gentzkow_et_al_2019} and the more recent \citet{Ash_Hansen_2023} provide great reviews of the different techniques that have been deployed to handle text in economics and finance research. For a recent survey of ``sentometrics'' (the econometrics of sentiment analysis) see \citet[][]{Algaba_et_al_2020}.}

\subsection{Lexical Approach}

The main approach to sentiment analysis in economics and finance has been the lexical methodology, which relies on a preset dictionary of words (a.k.a. \emph{lexicon}), with each word assigned a sentiment score.\footnote{Typically, words in the dictionary have a score of 1 if they are deemed to express positive sentiment, $-1$ if negative, and $0$ for neutral words. But more complex approaches exist.}

The basic input to any sentiment analysis task is a collection of $D$ text documents (the \emph{corpus}). The raw documents must first be converted into sequences of linguistic features (tokens). The lexical approach uses word-level tokenization almost exclusively.\footnote{A somewhat common exception is the use of short word sequences (2- and 3-grams) as units of analysis.} Therefore, document $d$ in the corpus can be written as $(w_{d,1}, \ldots, w_{d,N_d})$, where $N_d$ is the number of tokens found in the raw document.

The main challenge with the lexical approach is creating the dictionary. Without any adjustments, the vocabulary size---the number of distinct tokens in the corpus---can be a very large number. For this reason, the lexical approach typically involves some heavy pre-processing of the text. First, all connectors, punctuation and other \emph{stopwords} are eliminated from the documents. Second, words are lowercased and converted into their \emph{stem} form. These steps drastically reduce the dimensionality of the problem.

Within the lexical approach, the sentiment of a document is some simple aggregate of the scores of its constituent terms. The most common approach is to use the polarity score:
\begin{align*}
  \mathrm{polarity}_d &= \frac{\sum_i S^i_d}{\sum_i \abs{S^i_d}},
\end{align*}
where $S^i_d$ represents the sentiment score assigned by the lexicon to term $i$ of document $d$. For the simple case of $\pm1$ scores, this simplifies to
$$\frac{N^+_d - N^-_d}{N^+_d + N^-_d},$$
where $N^+_d$ and $N^-_d$ are, respectively, the counts of positive and negative terms in the document. The methodology is often referred to as ``bag of words'' because the order in which the terms appear does not affect the resulting sentiment score. Technically, however, bag of words is a broader concept, since there are machine learning approaches to sentiment extraction that ignore word-order without using a dictionary \citep[e.g.][]{Frankel_et_al_2022}.

The literature has devoted considerable attention to the quality and domain-specificity of the employed lexicons. For instance, \citet{Loughran_McDonald_2011} showed that the \emph{Harvard General Inquirer} (HGI, a commonly used dictionary) does not appropriately evaluate the sentiment in financial/economic text. Instead, they developed their own lexicon of terms for economics and finance, which to this day remains an important reference point.

Another attempt at improving on the lexicon approach adds a series of heuristic rules to take into account common linguistic patterns that affect sentiment, such as negation and emphasis. A prominent example is the \emph{Vader} analyzer \citep{Hutto_Gilbert_2014}, which is designed to deal with short texts such as tweets and online reviews.

Generally speaking, even the most sophisticated lexical approach does not take into account the \emph{context} in which words occur. It is not difficult to construct examples where several seemingly positive terms are used in a phrase in order to convey a strong negative opinion. Sarcasm, understatements, backhand comments, etc. are undetectable to lexicons.

In addition to this limitation, bag-of-words methods also suffer from the fact that the resulting sentiment score does not have an easy interpretation or unit of measurement. Moreover, while researchers customarily compare scores resulting from alternative lexicons, it is not in general possible to evaluate the relative accuracy of the different sentiment scores.

A final limitation of the lexical approach is that, if the method is to be used to generate a measure other than a sentiment score, a completely new dictionary must be produced. For example, \citet{Schmader_et_al_2007} created a set of dictionaries to summarize different semantic aspects of reference letters written for chemists and biochemists. Specifically, they developed sets of terms associated with the following language categories: grindstone traits, ability traits, research terms, teaching and citizenship terms, and standout adjectives. \citet{Eberhardt_et_al_2023} applied a similar strategy to letters of recommendation written on behalf of job market candidates applying to a top UK economics department.

\subsection{Fine-tuning LLMs for Sentiment Extraction}

Instead of attempting to extract numerical features directly (e.g. counting tokens present in a dictionary), language models represent text tokens with dense vectors called \emph{embeddings}, which are formed by having the model train on different language prediction tasks. The transformer architecture \citep{Vaswani_et_al_2017} revolutionized this approach by stacking attention layers on top of the embeddings in a way that allowed the model to learn to pick up important interactions across words in text sequences without the need to specify those dependencies explicitly. As a result, these models are able to encode the context-dependent nature of the meaning of text tokens.\footnote{See \citet{Zhao_et_al_2023} for a detailed review of the recent advances of LLMs.}

There are two main transformer variants, with the main contrast between them being the exact way in which the attention layers are applied. Encoder-only transformers use attention layers in which every token's embeddings are allowed to interact with every other token in the sequence. This type of model is pre-trained using \emph{masked language modeling}, a prediction task in which one or more tokens in a training sequence are masked (i.e. hidden from the model). The model's task is to predict the masked token or tokens using only the context as guidance. The great advantage of this form of training is that it is semi-unsupervised, meaning that there is no need for expensive labeled data. Instead, the model can be trained on massive corpora of raw text (e.g. Wikipedia).

Decoder-only (a.k.a. generative) transformers use ``causal'' attention layers, meaning that each token is only allowed to interact with the preceding tokens in the sequence.\footnote{There are other transformer variants (e.g. encoder-decoders for machine translation). I focus exclusively on the different approaches to sentiment analysis.} The pre-training task in this case involves predicting the correct continuation of truncated text sequences. This form of language modeling is also semi-unsupervised.

When transformers were first released, the state-of-the-art approach to using them was through a process called \emph{fine-tuning}. Starting with a pre-trained language model consisting of (hundreds of) millions of trained parameters, the researcher would add to the model a simple prediction head with randomly initialized weights. The type of prediction head would depend on the exact application. Since the vast majority of parameters in the model are already trained, the researcher would only need to ``fine-tune'' this model. This means training the final layer for just a few epochs with a very small learning rate (other model weights are frozen during fine-tuning) on a specially collected labeled data set. The idea is to minimally modify the model to adapt it to the desired task.

To put these points formally, consider first a supervised learning NLP task that takes a text $\boldsymbol x$ and predicts output based on a model $\mathrm{P}\left(y\mid \boldsymbol x; \boldsymbol \theta \right)$. For a text classification task, given an input text $\boldsymbol x$ the model predicts a label $y$ from a fixed label set $\mathcal{Y}$. For example, a sentiment analysis task might take an input such as ``I loved this movie!'' and predict a label $y=+$ out of the label set $\mathcal{Y}=\{+;\sim;-\}$. In this terms, the challenge of supervised learning is that in order to find optimal parameters, we need lots of supervised data $\left(\boldsymbol x_i, y_i\right),i=1,\ldots,N$ where the $y_i$ are the labels (a.k.a. ``ground truth'') for a number $N$ of training examples. Because language inputs are very high-dimensional\footnote{Roughly, the dimensionality of the input is the vocabulary size times the length of the longest sequence the model can consider.}, language models require millions (or billions) of parameters to achieve good performance. The larger the number of parameters, the more supervised data is required to train the model. This is the rationale behind the use of semi-unsupervised pre-training.

It is useful to partition the parameter vector as $\boldsymbol \theta\equiv \left[\boldsymbol \theta_{pre},\boldsymbol \theta_{out} \right]$, where $\boldsymbol \theta_{pre}$ comprises the embeddings and all other model weights except the output layer. In summary, the fine-tuning approach involves using versions of these weights that were obtained through language modeling (typically by an organization with a lot of computing power) and then training only $\boldsymbol \theta_{out}$ through supervised learning with domain-specific data. While the approach still requires some investment in labeling cases, the investment is typically negligible compared to what would be required without transferring the pre-trained weights.

One good example of fine-tuning for sentiment analysis is the \texttt{FLAIR} project \citep{Akbik_et_al_2019}.\footnote{See also \url{https://flairnlp.github.io/}.} Their model is based on \texttt{distilBERT}\footnote{The \texttt{distilBERT} \citep{Sanh_et_al_2020} model is an encoder-only transformer with a maximum context size of 512 tokens.} pre-trained weights. The output layer was fine-tuned using the Amazon review corpus and other similar text.

FinBERT \citep{Huang_et_al_2023} is another fine-tuned \texttt{BERT}-based model. Interestingly, the creators of \texttt{FinBERT} pre-trained the model themselves using a corpus of financial statements. They then fine-tuned it for sentiment analysis using labeled analyst reports.

The obvious limitation of fine-tuned models is their domain-specificity. Models like \texttt{FLAIR} or \texttt{FinBERT} cannot be expected to be good at predicting sentiment for documents outside the domain of their training data.

\subsection{Prompt-based Learning}

Soon after the first transformers were rolled out, it became apparent to researchers that the models had capabilities that far exceeded simple language generation. The newly discovered ``emergent properties'' of LLMs range from text classification to translation tasks \citep[see][]{Brown_et_al_2020}. As a result of this discovery, in recent years there has been a paradigm shift away from the fine-tuning approach and toward a \emph{prompt-based learning} approach.\footnote{For the sake of brevity, I limit the description of the paradigm shift and of prompt-based learning to the bare minimum. See \citet{Sun_et_al_2022}, \citet{Min_et_al_2023} and \citet{Liu_et_al_2023} for comprehensive reviews. \citet{Bu_et_al_2022} focus on sentiment analysis via prompt learning.}

Instead of fine-tuning an LLM to downstream tasks, these tasks can be reformulated to look more like those solved during the model's pre-training. The key insight is that a \emph{textual prompt} can be devised to elicit the desired prediction. For example, in order to recognize the emotion of the tweet ``I missed the bus today!'', we may append the prompt ``I felt so [MASK]'' and ask the LLM to fill the blank with an emotion-bearing word.\footnote{The same prompting strategy can be applied to multiple tasks. To elicit a translation, one might use the prompt: ``English: I missed the bus today! French: \ldots'' and use a generative LLM to generate the translation.}

The great advantage of reformulating the problem this way is that, once the prompting strategy has been set, one may proceed to making predictions \textbf{without the need of any supervised training}. Specifically, prompting spares researchers from the often unsurmountable task of collecting a large labeled data set.

Prompt-based learning takes advantage of the emergent properties of pre-trained model $\mathrm{P}\left(\boldsymbol x; \boldsymbol \theta_{pre} \right)$ of text $\boldsymbol x$ itself. For sentiment analysis, the derivation of predicted probabilities $\mathrm{P}\left(y\mid \boldsymbol x\right)$ requires three steps:

\paragraph{Step 1.} The first step involves the specification of a prompting function $f_{prompt}(\cdot)$. The prompting function takes text $\boldsymbol x$ as an input and produces a prompt: $\boldsymbol x^\prime = f_{prompt}(\boldsymbol x)$. The easiest way to specify $f_{prompt}(\cdot)$ is to use a template ($\mathcal{T}$), i.e. a text string that contains two slots: an input slot [X] for the input $\boldsymbol x$ and an answer slot [MASK] for an intermediate answer text $\boldsymbol m$.

For example, for the task of assigning sentiment to movie reviews, a suitable template would be: ``[X] Overall, it was a [MASK] movie.'' The resulting prompt would be
$$\boldsymbol x^\prime = \text{``I loved this movie! Overall, it was a [MASK] movie.''}$$

\paragraph{Step 2.} The second step of the prompting strategy involves defining a \emph{verbalizer} $\mathcal{V}(\boldsymbol m)$. Given a set of permissible answers $\mathcal{M}$, a verbalizer is just a mapping $\mathcal{V}(\boldsymbol m):\mathcal{M}\rightarrow\mathcal{Y}$ assigning each permissible answer to a label. In the movie review example, the permissible answers could be $\mathcal{M}= \{\text{``excellent''}; \text{``good''}; \text{``OK''};\text{``bad''}; \text{``horrible''}\}$. And the verbalizer could be defined as follows:

\begin{align*}
   \mathcal{V} =& \left\{(\text{``excellent''},+);(\text{``good''},+);(\text{``OK''},\sim); (\text{``bad''},-);(\text{``horrible''},-)\right\} \\
\end{align*}

\paragraph{Step 3.} The third and final step is simply to collect predicted probabilities for each label using the LLM. Let the function $f_{fill}(\boldsymbol x^\prime, \boldsymbol m)$  fill-in the location [MASK] in prompt $\boldsymbol x^\prime$ with the permissible answer $\boldsymbol m$. The predicted probabilities for each label are thus defined:

\begin{align}\label{eq_prob}
  \mathrm{P}(y\mid \boldsymbol x) =& \sum_{\mathbf{m}\in\mathcal{M}}\mathrm{I}\left[\mathcal{V}(\boldsymbol m)=y\right] \cdot
  \mathrm{P}\left( f_{fill}(f_{prompt}(\boldsymbol x), \boldsymbol m);\boldsymbol \theta_{pre} \right), \quad y\in \mathcal{Y}
\end{align}%
where $\mathrm{I}[\cdot]$ is the indicator function. Intuitively, the prompt-based probabilities are a measure of how likely the LLM judges each of the answered prompts. In particular, the predicted probability of assigning positive sentiment to the input text is simply the sum of the probabilities---based on the LLM's pre-training---assigned to each of the positive tokens in the verbalizer. This intuitive interpretation is one of the advantages of prompt-based sentiment extraction. In addition, because sentiment scores are derived from the predicted probabilities, they have a natural unit of measurement (p.p.).

The predicted probabilities can be used to predict a sentiment class $\hat{y}=\arg\max_{y\in\mathcal{Y}} \mathrm{P}(y\mid \boldsymbol x)$. However, in sentiment analysis it is customary to use the text's polarity\footnote{Polarity is a sentiment measure that is robust to over-sensitivity to subjective or otherwise sentiment carrying tokens in the input text.}, defined as:

\begin{align}\label{eq_polarity2}
  \mathrm{polarity}(\boldsymbol x) =& \mathrm{P}\left(y=+ \mid \boldsymbol x\right) - \mathrm{P}\left(y=-\mid \boldsymbol x \right)
\end{align}

In other words, prompt-based sentiment polarity is simply the difference between the sums of the probabilities that the model assigns to positive and negative completions of the prompt respectively. This intuitive interpretation greatly simplifies the use of prompt-based sentiment extraction for downstream tasks such as regression analysis.

\subsubsection{RLs: Sentiment Prompt and Verbalizer}

To extract sentiment from RLs, I use the template/verbalizer combination described in table~\ref{tabl_prompt_sentiment}.

\begin{table}[hbt!]
  \centering
    \caption{RL Prompt-based Sentiment Extraction}
    \label{tabl_prompt_sentiment}
    {\small
    \begin{tabular}{m{1.4in}p{0.4in}m{3.8in}}
      \toprule
      \multicolumn{1}{c}{\textbf{Prompt} ($\mathcal{T}_{sentiment}$)} & \multicolumn{2}{c}{\textbf{Verbalizer} ($\mathcal{V}_{sentiment}$)} \\
      \midrule
      ``[X] In summary, this job market candidate is [MASK]'' &&
      \begin{itemize}
      \item[$\textbf{positive}$:] \small{\{excellent, outstanding, exceptional, brilliant, great, superb, fantastic, terrific, amazing, phenomenal, awesome, stellar, wonderful, extraordinary, remarkable, impressive, incredible, fabulous, formidable, marvelous, perfect, fascinating\}}
      \item[$\textbf{negative}$:] \small{\{bad, terrible, disappointing, poor, awful, lousy, atrocious, pathetic, unsatisfactory, substandard, mediocre, common, inferior, average, ordinary, regular\}}
      \end{itemize} \\
      \bottomrule
    \end{tabular}
    }
\end{table}

\subsubsection{Model Choice}

In principle, any LLM capable of evaluating $\mathrm{P}\left(\boldsymbol{x};\boldsymbol{\theta_{pre}}\right)$ can be used for prompting. However, the model choice has important implications for the overall quality and precision of the sentiment scores, as well as for other somewhat more technical aspects:

\begin{enumerate}
  \item Generative models such as those in the \texttt{GPT} and \texttt{Llama} families require a \emph{prefix prompt}. In other words, the [MASK] term in the template must be the last token.\footnote{Because of this restriction, the [MASK] term is generally omitted altogether when presenting the prompt.} As mentioned above, the reason for this is that generative models are pre-trained using causal language modeling, i.e. predicting the next word in text sequences. In contrast, bidirectional models like \texttt{BERT}, \texttt{distilBERT}, \texttt{Big Bird}, or \texttt{ModernBert} are pre-trained via masked language modeling, so the [MASK] term can be anywhere in the template.\footnote{These are called \emph{cloze prompts}.}

  \item LLMs operate through a pipeline that begins with the \emph{tokenization} of the input text. Each model's tokenizer is somewhat different. The main restriction here is that the terms in the permissible set $\mathcal{M}$ must be in the tokenizer's vocabulary.\footnote{One important issue to check is whether multiple spellings of the same word are in the vocabulary. For example, both ``labor'' and ``labour'' are in most LLMs vocabularies.}

  \item An important consideration is that the length of the input text (measured in tokens) must fit within the model's \emph{context size}, which varies from model to model. If a document is longer than the context size, it will have to be processed using some kind of rolling-window approach.
\end{enumerate}

The main model I use to generate prompt-based sentiment scores is \texttt{Llama 3.1}, a large\footnote{I use the 8 billion parameter version, which is the largest that fits my GPU. There are larger versions, which probably would yield even better results.} generative language model pre-trained on a mix of publicly available online data ($\sim$15 trillion tokens). The model has a multi-lingual vocabulary encompassing 128,256 tokens. The context size---the maximum length of an input text sequence---is also 128k. For comparison purposes, in the online appendix I also present results based on prompting \texttt{ModernBert}, which is an encoder pre-trained on 1.7T tokens extracted from Wikipedia, the ``BookCorpus'', and code databases.\footnote{ModernBert is generally less capable than Llama: it has ``only'' 395M parameters, its vocabulary size is 50k tokens and its context size is 8,192. However, as already mentioned, the [MASK] token can be inserted anywhere when prompting a bidirectional model. In many applications, this flexibility makes creating an appropriate prompt much easier, so it is interesting to see if a bidirectional model attains reasonably comparable performance.}

\subsubsection{Foundational v. Instruct Models}

All the LLMs mentioned so far belong to the \emph{foundational} class, meaning that they have not been fine-tuned to follow human instructions or to produce output that is ``aligned'' with the objectives of humans. \emph{Instruct} models, like \texttt{Llama-instruct} \citep{Touvron_et_al_2023}, \texttt{GPT-instruct} \citep{Ouyang_et_al_2022}, or \texttt{DeepSeek-V3} \citep{deepseekai_2025} are ``post''-trained using supervised data of question-answer pairs, as well as other techniques like reinforcement learning with human feedback, in which the model produces several alternative outputs that a human user later ranks.\footnote{World famous applications like ChatGPT make further additions. For example, all chatbots incorporate a memory layer to the otherwise stateless instruct model.}

An important additional difference is that it is not possible to access the complete output---i.e. the complete probability distribution over the vocabulary---of instruct models. Instead, the model generates text output based on a sampling procedure that uses the output probabilities as weighting factors.\footnote{There are additional parameters that inform the sampling process, like the `temperature' parameter.} While it is possible to get the probabilities associated with generated tokens from the instruct model's output, it is \textbf{not} in general possible to get the probabilities of all the tokens that were not sampled. Because of this, it is not possible to calculate a polarity score such as equation~\eqref{eq_polarity2}.

Despite this loss of information, there are advantages to using instruct-type models in order to obtain the most likely label for the text input. These models allow very sophisticated prompting functions, most notably the ability to assume different ``roles'' and to format the model's answers following the user's specifications. For comparison purposes, I obtain sentiment scores from \texttt{Llama 4 Instruct} and \texttt{DeepSeek-V3} using the following instructions:

\begin{samepage}
    \begin{center}\small{
        \begin{tabular}{lp{11cm}}
          System Prompt & ``You are a helpful assistant. Analyze the letter of recommendation submitted by the user and assess how positive the letter writer is regarding the ability level and future prospects of the job market candidate. Take into account that letter writers tend to be overly optimistic regarding job market candidates. Only give the maximum score to candidates deemed exceptional. Most candidates should get a medium score (around 0).''\\
          Human Prompt  & ``Below is a letter of recommendation for a job market candidate: [X]''\\
          Format Instructions & The only admissible output is a floating point number between -1 and 1.\\
        \end{tabular}
        }
    \end{center}
\end{samepage}

It is important to reiterate that the ``score'' that results from these instructions is quite unlike the polarity score from equation~\eqref{eq_polarity2}. The former consists of a sequence of \emph{text tokens} denoting a floating point number within -1 and 1 that results from the instruct model's sampling process. The latter is a \emph{numerical} outcome resulting from summarizing the foundational model's output vector of probabilities.

An unfortunate implication of the sampling process inherent in text generation by instruct-based LLMs is that it can be difficult to ensure perfect reproducibility of results. Modern APIs allow for setting a seed number, which should in principle ensure reproducibility as long as other parameters (prompt, temperature, model version, etc.) remain unchanged. However, inference providers often update their platforms and this may on occasion affect reproducibility even for same-seed constant-prompt generation. This challenge is altogether absent when using foundational LLMs' output in a deterministic way.

\section{Hiring Data}

The analysis in this paper is based on administrative data from a leading research department in Russia during the 2013, and 2015--2021 job markets.\footnote{I designate job market cycles by the year in which recruitment formally started and the job ads were posted. Unfortunately, data for the 2014 market was not archived.} To protect confidentiality, I processed all RL text data myself.

As already mentioned, I only consider job market candidates who are finishing their PhD programs and have not held a full-time academic position. Overall, my data contains 1,968 letters from 645 job market candidates. In some of the analysis, I use a sample of ``complete applications'' consisting of candidates for whom I have at least three letters, including the letter from the main research adviser. This latter sample has 1,800 letters from 553 candidates.

The pre-processing of the reference letters' text was kept to a minimum. I removed headers, footers, page numbers, salutations, and some obvious mistakes from faulty conversions from pdf format. I also collected personal data from the submitted CVs, as well as personal websites and LinkedIn profiles. Specifically, I have data on the candidate's sex, nationality, prior publications and time to PhD completion. I also collect the PhD granting institution's location and ranking.\footnote{To rank Economics departments, I use Tilburg University's ranking, which is based on publications weighted by journal impact factor and covers the whole world (\url{https://rankings.tilburguniversity.edu/}).} Finally, I hand-collected the detailed job market outcome for each candidate in my data. The job market outcome corresponds to the job held by September of the following year.

In addition, I collected data on the characteristics of the letter writers: sex, affiliation, formal position, experience (years from PhD completion), and publication rank.\footnote{I focus on whether the letter writer is in the top-10 or top-5 percentile of RePec's publications ranking (\url{https://ideas.repec.org/top/top.person.alldetail.html}).}

\subsection{Descriptive Statistics}

\begin{table}
  \centering
  \caption{Descriptive Statistics --- Job Market Candidates}\label{tabl_desc_cand}
  {\scriptsize
  \begin{tabular}{p{1.2in}ccc|p{1.2in}ccc}
\toprule
    & Mean & Min & Max  & & Mean & Min & Max\\
\midrule
\multicolumn{4}{l|}{\textbf{Gender}}          & \multicolumn{4}{l}{\textbf{Job Market Year}} \\
 Female & 0.264 & 0 & 1                       & 2013 &	0.118   & 0 & 1\\
\multicolumn{4}{l|}{\textbf{Region of Origin}}& 2015 & 0.101 & 0 & 1\\
  Developing Country   & 0.093 & 0 & 1        & 2016 & 0.098 & 0 & 1\\
  Former Soviet Union  & 0.209 & 0 & 1        & 2017 & 0.091 & 0 & 1\\
  North America  & 0.040 & 0 & 1              & 2018 & 0.127 & 0 & 1\\
  Europe  & 0.354 & 0 & 1                     & 2019 & 0.172 & 0 & 1\\
  Asia  & 0.305 & 0 & 1                       & 2020 & 0.149 & 0 & 1\\
\multicolumn{4}{l|}{\textbf{Research Field}}  & 2021 & 0.144 & 0 & 1\\
  Finance  & 0.271 & 0 & 1               & \multicolumn{4}{l}{\textbf{Number of RLs}} \\
  Applied  & 0.341 & 0 & 1               & 1  & 0.028  &  0  & 1  \\
  Macroeconomics  & 0.236 & 0 & 1        & 2  & 0.112  &  0  & 1 \\
  Theory  & 0.098 & 0 & 1                & 3  & 0.642  &  0  & 1  \\
  Econometrics  & 0.054 & 0 & 1          & 4  & 0.219  &  0  & 1  \\
\multicolumn{4}{l|}{\textbf{PhD Rank}}    & \multicolumn{4}{l}{\textbf{PhD Region}}   \\
 1--25  & 0.251 & 0 & 1                  & USA  & 0.493 & 0 & 1    \\
 26--50  & 0.229 & 0 & 1                 & \multicolumn{4}{l}{\textbf{PhD Length}} \\
 51--75  & 0.087 & 0 & 1                 & Seven years or more  & 0.096 & 0 & 1  \\
 76--100  & 0.073 & 0 & 1                & \multicolumn{4}{l}{\textbf{Publications}} \\
 101--150  & 0.144 & 0 & 1               & At Least One Top Field  & 0.053 & 0 & 1 \\
 151--200  & 0.079 & 0 & 1               & \multicolumn{4}{l}{}\\
 201+  & 0.136 & 0 & 1                   & \multicolumn{4}{l}{}\\
\bottomrule
\multicolumn{8}{p{4.9in}}{Notes: $N$=645. The PhD ranking is from Tilburg University. I considered publications in JIE, JET, JoE, JME, JPubE, JEH, JFE, JF, Rand, JEEA, REStat, EJ, IER, all the AEJs, and the top 5.} \\
\end{tabular}

  }
\end{table}

\begin{table}
  \centering
  \caption{Descriptive Statistics --- Letter Writers}\label{tabl_desc_writers}
  {\scriptsize
  
\begin{tabular}{lcccc}
    \toprule
     &   N & Mean & Min & Max \\
    \midrule
    \multicolumn{5}{l}{\textbf{Title}} \\
     Full Professor      & 1968 & 0.609 & 0 & 1 \\
     Associate Professor & 1968 & 0.208 & 0 & 1 \\
     Assistant Professor & 1968 & 0.158 & 0 & 1 \\
     Non-academic        & 1968 & 0.025 & 0 & 1 \\
    \midrule
    \textbf{Female}                         & 1968 & 0.119 & 0 & 1 \\
    \textbf{Experience} (years since PhD granted) & 1968 & 18.35 & 0 & 56 \\
    Ranked in Repec \textbf{Top 10\%} & 1968 & 0.511 & 0 & 1 \\
    Ranked in Repec \textbf{Top  5\%} & 1968 & 0.345 & 0 & 1 \\
    \textbf{Main Research Adviser}     & 1968 & 0.310 & 0 & 1 \\
    \textbf{External Referee} & 1968 & 0.194 & 0 & 1 \\
    \textbf{Ref. from other field} & 1968 & 0.360 & 0 & 1 \\
    \bottomrule
    \multicolumn{5}{p{3.6in}}{Notes: There are 1,578 unique letter writers: 1,303 wrote a single letter and 275 wrote multiple letters. `External' referees are not affiliated with the candidate's university.} \\
\end{tabular}

  }
\end{table}

Table~\ref{tabl_desc_cand} summarizes the main characteristics of the job market candidates in my data. There are some notable differences between my data and similar samples collected by \citet{Eberhardt_et_al_2023} and \citet{Baltrunaite_et_al_2024} from British and Italian institutions, respectively. Given that my institution is based in Russia, it is not surprising to see an over-representation of applicants from the former Soviet Union.\footnote{Of these, 108 are Russian, 8 are Ukrainian, 5 are Armenian, 5 from the Baltic Countries, and 5 from other countries. There is also a relatively small fraction of North Americans.} Just over 26\% of applicants in my data are female, compared to around 30\% in their samples. Another important difference is that the samples in their studies include both new PhDs and experienced academics. Because of this, the candidates in their samples have more publications.

In other respects, the samples are comparable. In particular, the fraction of candidates getting their PhD from an American institution (around 50\%) and the fraction in top-ranked programs are very similar.

Table~\ref{tabl_desc_writers} presents some statistics regarding the letter writers. The main observation here involves how senior most letter writers are. Over 60\% of the RL were written by someone with the rank of full-professor. The average writer has over 18 years of potential experience and almost 35\% of writers are in the top-5th percentile in RePec's ranking of authors in economics. A second important point is that only about 12\% of the writers are women. These statistics are roughly comparable to those found in other studies with similarly collected samples.

\subsection{Reference Letter Sentiment}

\begin{figure}[htb!]
    \centering
      \caption{The Distribution of Prompt-based RL Sentiment}
      \label{fig_scatter_Llama31}
    \begin{subfigure}[b]{0.475\textwidth}
        \includegraphics[scale=0.8]{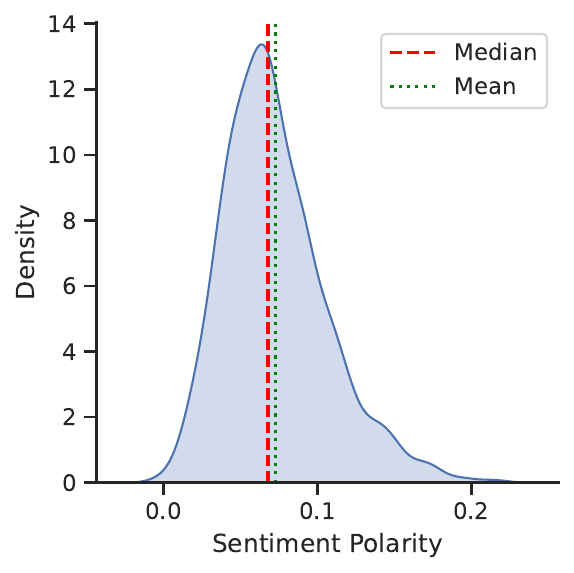}
    \end{subfigure}
    \hfill 
    \begin{subfigure}[b]{0.475\textwidth}
        \includegraphics[scale=0.8]{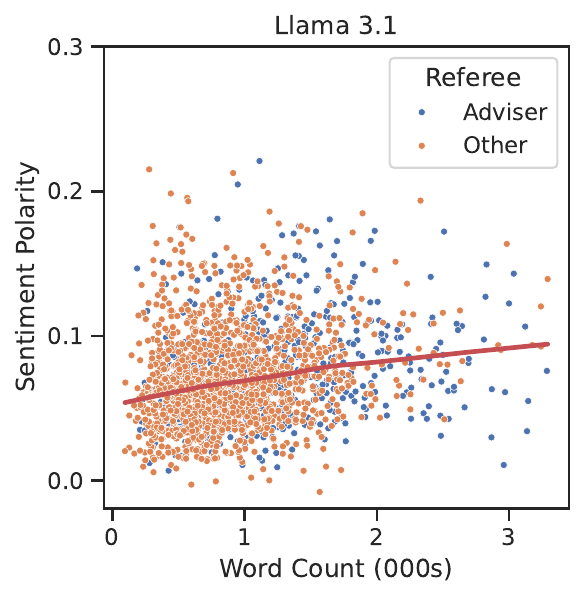}
    \end{subfigure}
    \vspace{-0.175in}
    {\small \flushleft \hspace{-1.65in} Notes: $N$=1968. Fitted line is a locally weighted regression.}
\end{figure}

Figure~\ref{fig_scatter_Llama31} (left) shows the kernel estimate of the distribution of prompt-based polarity scores for the full sample of RLs. Several facts are apparent. Sentiment polarity is overwhelmingly positive; there are only a couple of letters with negative polarity. The distribution of sentiment is roughly symmetric, with mean sentiment of about 0.07.\footnote{Table~\ref{tabl_sentiment_descriptives} in the appendix has descriptive statistics for all the sentiment measures used in this paper.}

There is a clear positive relationship between letter length and letter sentiment (right panel). Advisers' letters are distinctly longer than other referees' letters (on average, almost 33\% longer). However, there is no significant ``adviser effect,'' i.e. there is no significant difference between the adviser's and other writer's sentiment independent of letter length. Frequent RL readers will not be surprised by any of these points.

\begin{figure}[htb!]
  \centering
  \caption{Prompt-based Sentiment Polarity Covariates}
  \label{fig_sent_by_categ}
      \begin{subfigure}{0.38\textwidth}
         \caption{\textbf{Gender}}
         \includegraphics[scale=0.7]{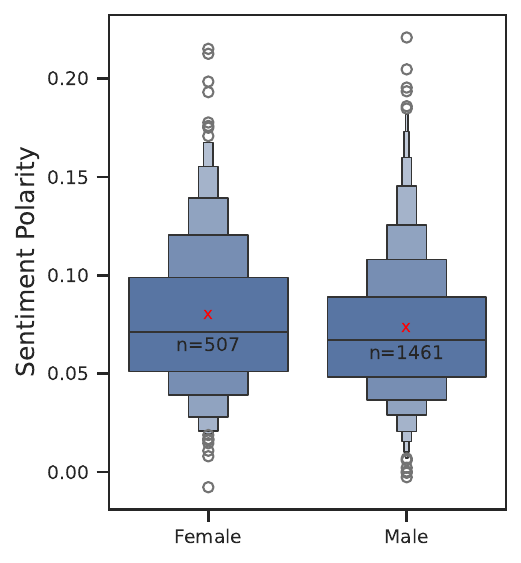}
      \end{subfigure}
      \begin{subfigure}{0.58\textwidth}
         \caption{\textbf{Candidate Region}}
         \includegraphics[scale=0.7]{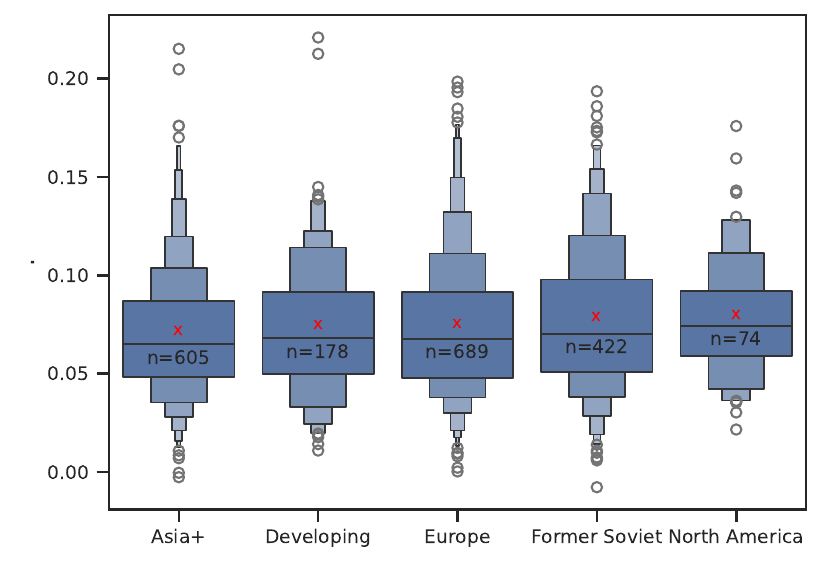}
      \end{subfigure}
      \begin{subfigure}{0.38\textwidth}
         \caption{\textbf{PhD Region}}
         \includegraphics[scale=0.7]{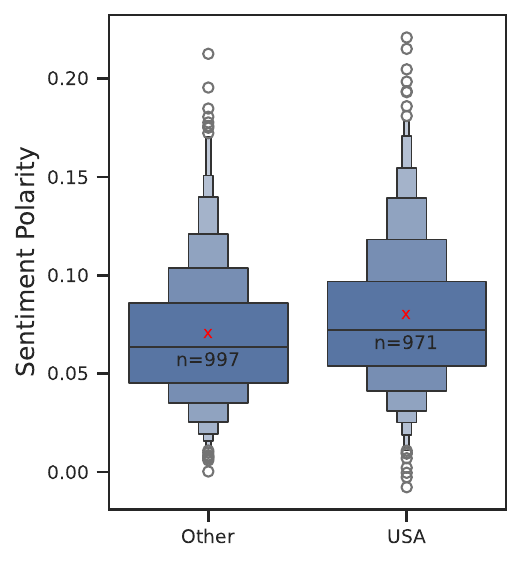}
      \end{subfigure}
      \begin{subfigure}{0.58\textwidth}
         \caption{\textbf{PhD Rank}}
         \includegraphics[scale=0.7]{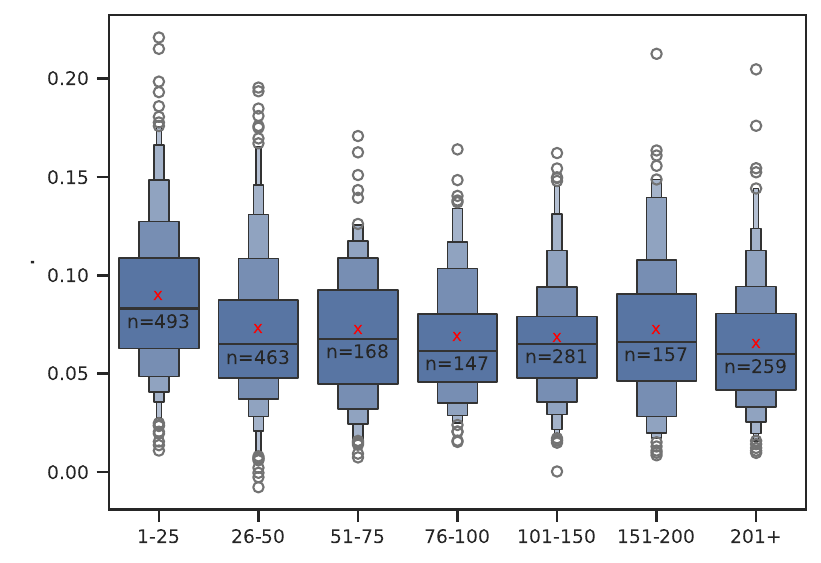}
      \end{subfigure}
      \begin{subfigure}{0.38\textwidth}
         \caption{\textbf{Publications}}
         \includegraphics[scale=0.7]{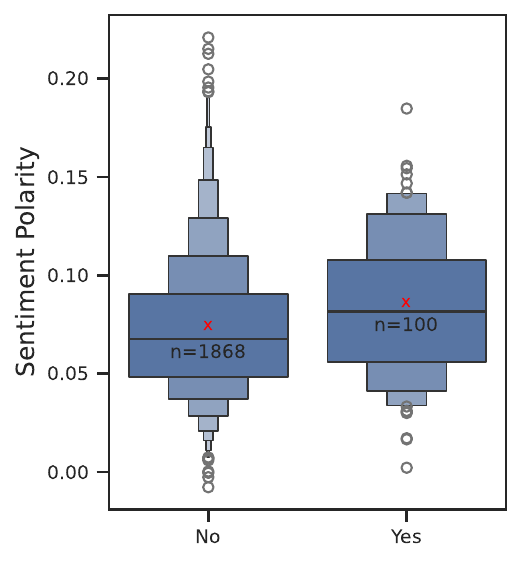}
      \end{subfigure}
      \begin{subfigure}{0.58\textwidth}
         \caption{\textbf{Letter Writer's Rank}}
         \includegraphics[scale=0.7]{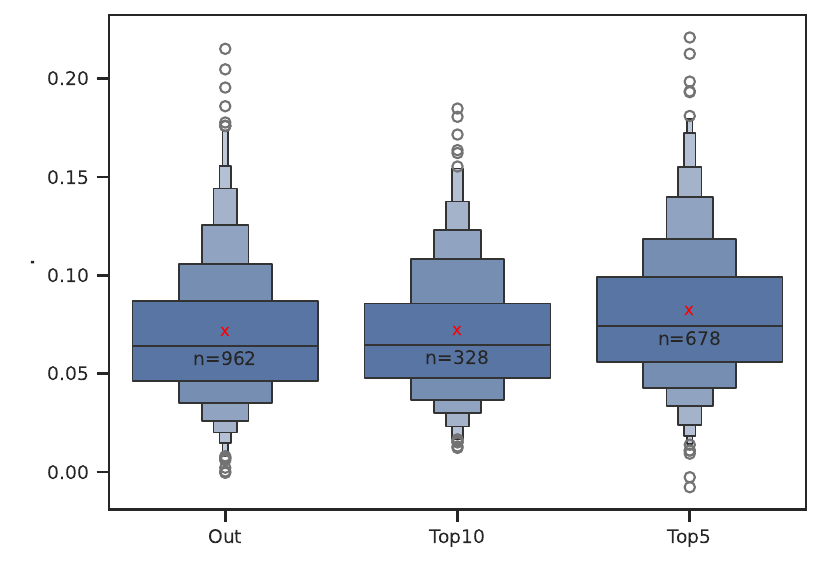}
      \end{subfigure}
    \vspace{-0.15in}
    {\small \flushleft \hspace{-0.8in} Notes: Horizontal lines mark the median score. Red crosses mark the mean.}
\end{figure}

Figure~\ref{fig_sent_by_categ} shows how the distribution of prompt-based sentiment polarity varies by different covariates. I focus on categories where differences are strongly statistically significant. I find a small difference in average sentiment scores in favor of women and against candidates from Asia and from Developing countries.

Letter sentiment tends to reinforce other known signals of higher ability. Candidates from PhD programs in America and from programs in the top 25 have substantially better RLs. The same is true about candidates with top publications, as well as candidates who managed to procure a letter from a top-5 percentile research economist. These statistics are in line with what a general sentiment measure should look like under the general assumption that higher ability candidates get more enthusiastic references.

\subsection{Testing Prompt-based Learning}

While it is possible to generate labels to evaluate a sentiment-scoring model, doing so is costly and not necessarily convincing since it would be hard to hire reviewers with subjective opinions similar to those of hiring committees at research economics departments. Moreover, because of confidentiality and other considerations, it is particularly hard to label RLs.

Nevertheless, it is still possible to test the prompt-based procedure by predicting readily available meta-data that one can safely assume is contained in most RLs. Here I provide an example consisting in predicting the sex and research field of the job market candidate described in the letter. For this purpose, I addressed the model with the prompts and verbalizers detailed in table~\ref{tabl_testing_prompt}.

\begin{table}[htb!]
  \centering
    \caption{Tests of Prompt-based Learning}
    \label{tabl_testing_prompt}
    {\small
    \begin{tabular}{m{1.7cm}p{0.3cm}m{3.5cm}p{0.3cm}m{8cm}}
        \toprule
  \multicolumn{1}{c}{\textbf{Test}}& \multicolumn{2}{c}{\textbf{Prompt} ($\mathcal{T}$)}& \multicolumn{2}{c}{\textbf{Verbalizer} ($\mathcal{V}$)} \\
  \midrule
      \textbf{Candidate Sex}   &&``[X] The job market candidate is a young [MASK]'' && \vspace{-0.4cm}
        \begin{itemize}
          \item[$\textbf{male}$:] $\{$man, male, gent$\}$
          \item[$\textbf{female}$:] $\{$woman, female, lady$\}$
        \end{itemize} \\
        \midrule
      \textbf{Field of Research} &&  ``[X] This job market candidate is an expert in the field of [MASK]'' &&
      \begin{itemize}
        \item[$\textbf{applied}$:] $\{$labor, labour, industrial, population, family, education, health, development, experimental, household, urban, resource, ecological, environmental, agricultural, transition, comparative, culture, cultural, behavioral$\}$
        \item[$\textbf{macro}$:] $\{$macro, monetary, mac, global, growth, fiscal, government, dynamic$\}$
        \item[$\textbf{finance}$:] $\{$finance, financial, business, asset, managerial, risk, portfolio, corporate$\}$
        \item[$\textbf{theory}$:] $\{$micro, mathematics, mathematical, cooperative, mechanism, decision, choice, strategic, auction, auctions, theory, theoretical, network, game, games, information$\}$
        \item[$\textbf{metrics}$:] $\{$statistical, econ, bayesian, stochastic, computational, instrument, instrumental, structural, inference$\}$
      \end{itemize} \\
      \bottomrule
    \end{tabular}
    }
\end{table}

\begin{table}
  \centering
  \caption{Prompt-based Learning Performance}\label{tabl_prediction_results}
  {\scriptsize
  
\begin{tabular}{lcccc}
    \toprule
\multicolumn{1}{p{1.3in}}{A.~\textbf{Candidate Sex}}& Precision  & Recall  & F1-score & Support      \\ \midrule
      Female &     1.00  &   0.99  &   1.00   &   507   \\
        Male &     1.00  &   1.00  &   1.00  &   1461    \\
        \midrule
    Accuracy &           &         &   1.00  &   1968    \\
   Macro Avg.&      1.00 &    1.00 &   1.00  &   1968    \\
Weighted Avg.&      1.00 &    1.00 &   1.00  &   1968    \\[0.1cm]
    \toprule
\multicolumn{1}{p{1.3in}}{B.~\textbf{Field of Research}}& Precision  & Recall  & F1-score & Support      \\ \midrule
  Applied       & 0.83   &  0.82  &   0.82   &   673  \\
  Econometrics  & 0.73   &  0.85  &   0.79   &   108  \\
  Finance       & 0.94   &  0.87  &   0.90   &   532  \\
  Macroeconomics& 0.79   &  0.85  &   0.82   &   466  \\
  Theory        & 0.75   &  0.74  &   0.74   &   189  \\
  \midrule
  Accuracy      &        &        &   0.83   &  1968  \\
  Macro Avg.    &  0.81  &   0.83 &   0.82   &  1968  \\
  Weighted Avg. &  0.84  &   0.83 &   0.83   &  1968  \\
    \bottomrule
\end{tabular}

  }
\end{table}

Table~\ref{tabl_prediction_results} shows the model's predictive performance for these tasks. The sex test is admittedly simple and could probably be solved by a dictionary counting pronouns as well. Basically, prompt-based learning gets it right in all but four cases.

Predicting field of research is not an easy task and it would probably have high error rates even if a human read and labeled each RL. First, not every RL really mentions the field of research of the candidate. There are ``teaching'' references and other short letters that do not really delve into the candidate's research. Second, lots of candidates do work on multiple fields, or fields that are not pure cases (e.g. `applied theory', 'central banking', `macro labor', `structural econometrics', etc.). In sum, classifying candidates into fields based on a RL is far from trivial. Given these considerations, prompt-based learning performs remarkably well, with accuracy and average F-1 scores well above 80\%.

\section{Predicting Job Market Outcomes}

\begin{figure}[htb!]
  \centering
  \caption{Job Market Outcomes}\label{fig_outcomes}
         \includegraphics[scale=0.52]{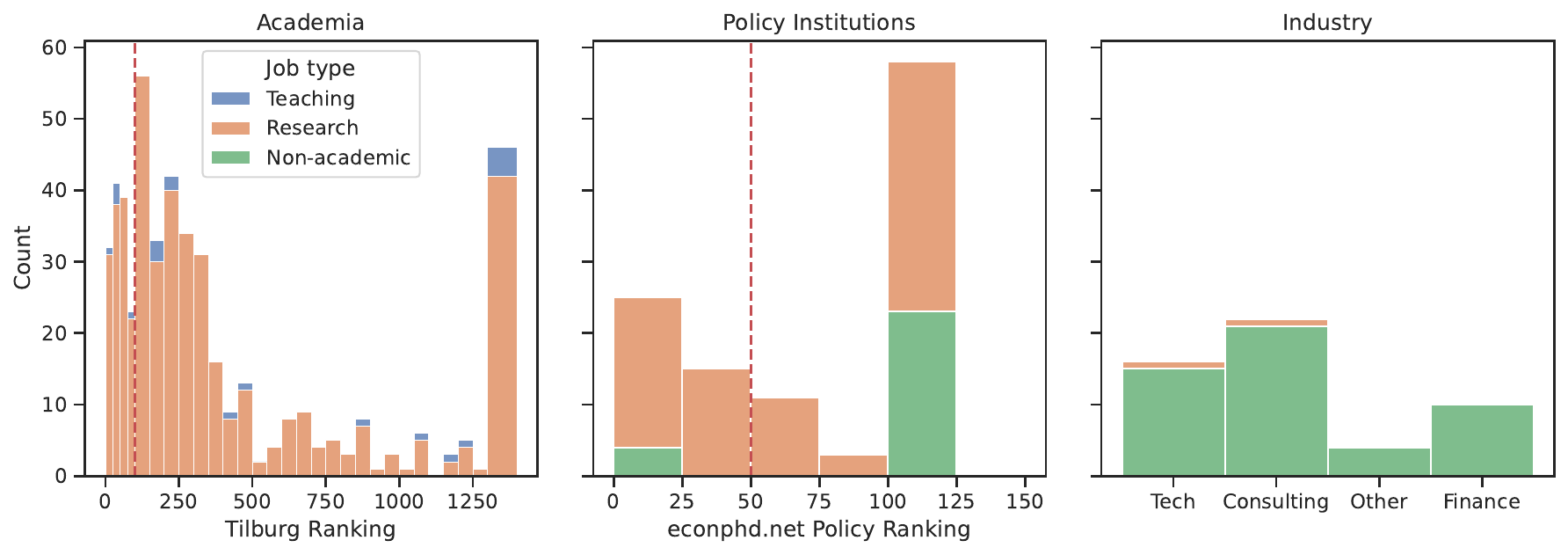}
    \vspace{-0.2in}
    {\small \flushleft \hspace{-0.38in} Notes: $N$=645. Policy institutions include central banks, development agencies and think tanks.}
\end{figure}

There are multiple potentially appropriate measures of what constitutes ``success'' in the academic job market. Job market outcomes are heterogenous on multiple fronts. Consider figure~\ref{fig_outcomes}, depicting different aspects of candidates' placement by September after the job market season. Outcomes vary by the sector of employment and the type of position. In addition, employers differ on how they are ranked according to their research output, with higher ranked employers being markedly more desirable to the average junior academic.\footnote{In order to determine the employer's ranking, I once again use Tilburg's ranking for academic institutions. For policy institutions, I use the \texttt{econphd.net} ranking (\url{https://econphd.econwiki.com/rank/rallec.htm}), which focuses on publications and has a transparent methodology.}

\vspace{0.2in}

In what follows, I define my baseline ``success'' outcome as attaining a top-100 academic or top-50 policy job. For some clarity on these cutoffs, consider that economics departments ranked 98--102 in the Tilburg ranking are NHH (Norwegian School of Economics), INSEAD, the California Institute of Technology and the University of Gothenburg. Institutions ranked in the top-50 of the policy ranking include the Federal Reserve Banks and the Board of Governors, as well as the ECB and a few other European national banks, plus the World Bank, the IMF and a couple other major development institutes. In figure~\ref{fig_outcomes}, the baseline cutoffs are depicted by vertical red lines. Clearly, only a minority of candidates in my sample attained success based on this measure.

As a robustness check, I also define four alternative outcome variables. The main modification in alternative~1 is to exclude post-docs from the definition of success, which arguably are less desirable than tenure-track positions even if they are sponsored by top universities. Alternative~2 considers the possibility that some positions at tech companies are as desirable as top academic placements. Finally, alternatives 3 and 4 simple change the cutoffs for success. The full definitions of these variables, as well as basic descriptive statistics, can be found in table~\ref{tabl_depvars_def}.

\begin{table}[htb!]
  \centering
  \caption{Job Market Success: baseline and alternative definitions}\label{tabl_depvars_def}
    {\scriptsize
        \begin{tabular}{lm{8cm}m{1cm}m{1cm}}
    \toprule
\multicolumn{1}{c}{\textbf{Outcome}}&\multicolumn{1}{c}{\textbf{Definition}}&\multicolumn{1}{c}{$\mathbf{N}_{success}$}&\multicolumn{1}{c}{$\mathbf{\%}_{success}$}\\
    \midrule
    Baseline & Top-100 Academic Employers \& Top-50 Policy Institutions & 175 & 27.1 \\[0.2cm]
Alt.~1 & Exclude teaching positions (liberal arts colleges) and post-docs from the success group& 115& 17.8 \\[0.2cm]
    Alt.~2 & Add some highly paid tech jobs (Amazon, Meta, etc.) to the success group & 190 & 29.5 \\[0.2cm]
    Alt.~3 & Use somewhat more forgiving cutoffs: top-150 academic and top-75 policy & 243 & 37.7 \\[0.2cm]
    Alt.~4 & Use somewhat more restrictive cutoffs: top-75 academic and top-35 policy & 150 & 23.3 \\
    \bottomrule
    \multicolumn{4}{l}{Notes: $N$=645.} \\
    \end{tabular}

    }
\end{table}

\begin{figure}[htb!]
  \centering
  \caption{Average RL Sentiment and Success in the Academic Job Market}\label{fig_sent_by_success}
         \includegraphics[scale=0.8]{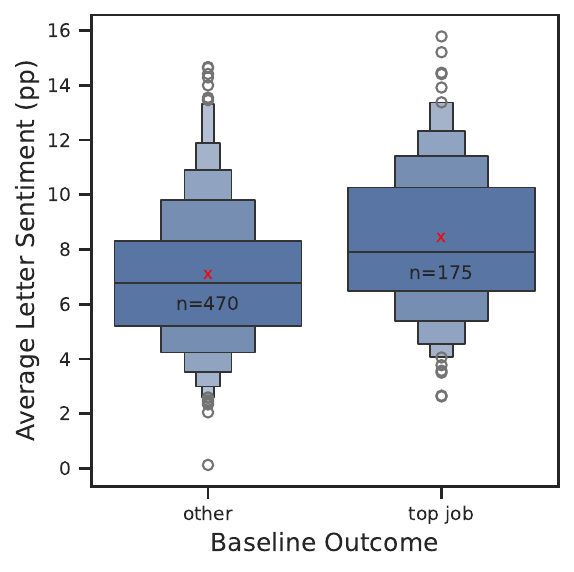}
    \vspace{-0.15in}
    {\scriptsize \flushleft 
     \begin{tabular}{p{2.85in}}
       Notes: The difference in means is $1.33$~p.p. The HC3 standard error is~$0.23$. \\
     \end{tabular}
     }
\end{figure}
\vspace{0.2in}

Figure~\ref{fig_sent_by_success} depicts the relationship between average RL sentiment and baseline success in the academic job market. Clearly, candidates who succeeded to obtain a top job have letters with significantly higher average sentiment.

\subsection{Linear Regressions}\label{section_OLS}

In order to quantify the relationship between the sentiment contents of a candidate's RLs and their job market outcome, I first look at the regression model:

\begin{align}\label{eq_OLS_avg}
 y_{it} =& \:\alpha+\beta\,S_{it}+\boldsymbol{X}^{\prime}_{it}\boldsymbol\gamma+\epsilon_{it}
\end{align}
where $y_{it}$ is the job market outcome of candidate $i$, in job market year $t$. The main variable of interest in this model is $S_{it}$, the average sentiment polarity (in p.p.) of the RLs submitted on behalf of the candidate.\footnote{$S_{it}=\sum_{l=1}^{L_i}S_{it}^l/L_i$, where $S_{it}^l$ are the individual RL sentiment scores and $L_i$ is the number of RLs submitted on behalf of candidate $i$.} All regressions include controls for the number of RLs and the average RL length (in thousands of words). Depending on the exact specification, the vector of control variables may also include: the candidate's individual characteristics (indicators for gender, region of origin, field of research, publications, and a long PhD), the PhD program's characteristics (department rank and region), the job market year, and whether the letter writers are senior scholars (specifically, the number of RePec top-5 percentile letter writers).

RLs are only one input in the recruitment process. Job market outcomes also depend on how the candidates present themselves in interviews, seminar presentations, and one-on-one meetings with faculty and administrators at potential employers, among other variables. Because these other determinants remain unobserved, the regression coefficients cannot be interpreted causally.\footnote{Assuming strong candidates get good RLs and also tend to perform well in interviews and seminars, one reasonable expectation would be for regression coefficients to \emph{over-estimate} the true causal effect that RL sentiment might have on a candidate's probability of success.}

However, in light of the theoretical literature regarding whether referees can be trusted to offer honest assessments and whether hiring committees are properly incentivized to select the best candidates, it is still interesting to determine whether a statistically significant association exists between good letters and good job market outcomes. In this section and the next, I show that not only are higher average sentiment RLs predictive of better outcomes, but also that the degree of disagreement among the letter writers---what I call \emph{sentiment dispersion}---is negatively associated with success in the market.

\begin{table}[tb!]
  \centering
  \caption{Predicting Success in the Academic Job Market}\label{tabl_OLS_baseline_Llama_avg}
  {\scriptsize
  \begin{tabular}{lccccc}
\toprule
  & \multicolumn{5}{c}{\emph{Baseline Outcome}: top-100 academic and top-50 policy jobs} \\
 & (1) & (2) & (3) & (4) & (5) \\
\midrule
 Average RL Sentiment & 0.0304$^{\phantom{***}}$ & 0.0276$^{\phantom{***}}$ & 0.0230$^{\phantom{***}}$ & 0.0233$^{\phantom{***}}$ & 0.0227$^{\phantom{***}}$ \\
  (p.p.)    & (0.0072)$^{***}$ & (0.0070)$^{***}$ & (0.0072)$^{***}$ & (0.0074)$^{***}$ & (0.0074)$^{***}$ \\
            & [0.0067]$^{***}$ & [0.0062]$^{***}$ & [0.0063]$^{***}$ & [0.0064]$^{***}$ & [0.0064]$^{***}$ \\
  & $|$0.0079$|^{***}$ & $|$0.0078$|^{***}$ & $|$0.0092$|^{**\phantom{*}}$ & $|$0.0096$|^{**\phantom{*}}$ & $|$0.0098$|^{**\phantom{*}}$ \\
\midrule
\multicolumn{1}{l}{RL Count \& Avg. Length}     & Yes& Yes & Yes & Yes & Yes \\
\multicolumn{1}{l}{Candidate's characteristics} & No & Yes & Yes & Yes & Yes \\
\multicolumn{1}{l}{PhD characteristics}         & No & No  & Yes & Yes & Yes \\
\multicolumn{1}{l}{Job Market Year}             & No & No  & No  & Yes & Yes \\
\multicolumn{1}{l}{Number of top-5\% Writers}     & No & No  & No  & No  & Yes \\
 Adjusted $R^2$ & 0.089 & 0.135 & 0.141 & 0.132 & 0.130 \\
\bottomrule
\multicolumn{6}{p{5.1in}}{Notes: $N=645$. Standard Errors: (Independent: HC3) [Clustered: Candidate Univ] $|$Clustered: Univ Rank$\times$Period$|$. $^{**}$p$<$0.05; $^{***}$p$<$0.01. Candidate's characteristics: sex, region of origin, field of research, major publications, PhD lasting 7 years or longer. PhD program characteristics: rank, region. See appendix table~\ref{tabl_OLS_baseline_Llama_avg_controls} for control variable estimates.} \\
\end{tabular}

  }
\end{table}

\begin{table}[tb!]
  \centering
  \caption{Alternative Definitions of Success}\label{tabl_OLS_robust_Llama_avg}
  {\scriptsize
  
\begin{tabular}{lccccc}
\toprule
 \multicolumn{1}{c}{\emph{Outcomes}$\rightarrow$}      & \multicolumn{1}{c}{\emph{Baseline}$^{\phantom{***}}$} & \multicolumn{1}{c}{\emph{Alt. 1}$^{\phantom{***}}$} & \multicolumn{1}{c}{\emph{Alt. 2}$^{\phantom{***}}$} & \multicolumn{1}{c}{\emph{Alt. 3}$^{\phantom{***}}$} & \multicolumn{1}{c}{\emph{Alt. 4}$^{\phantom{***}}$} \\
\midrule
 Average Letter Sentiment & 0.0227$^{\phantom{***}}$ & 0.0147$^{\phantom{***}}$ & 0.0237$^{\phantom{***}}$ & 0.0276$^{\phantom{***}}$ & 0.0229$^{\phantom{***}}$ \\
 (p.p.)                   & (0.0074)$^{***}$ & (0.0064)$^{**\phantom{*}}$ & (0.0077)$^{***}$ & (0.0086)$^{***}$ & (0.0073)$^{***}$ \\
        & [0.0064]$^{***}$ & [0.0057]$^{***}$ & [0.0068]$^{***}$ & [0.0081]$^{***}$ & [0.0064]$^{***}$ \\
        & $|$0.0098$|^{**\phantom{*}}$  & $|$0.0072$|^{**\phantom{*}}$  & $|$0.0088$|^{***}$  & $|$0.0098$|^{***}$ & $|$0.0093$|^{**\phantom{*}}$ \\
\midrule
\multicolumn{1}{l}{RL Count \& Avg. RL Length}     & Yes& Yes & Yes & Yes & Yes \\
\multicolumn{1}{l}{Candidate's characteristics} & Yes& Yes & Yes & Yes & Yes \\
\multicolumn{1}{l}{PhD characteristics}         & Yes& Yes & Yes & Yes & Yes \\
\multicolumn{1}{l}{Job Market Year}             & Yes& Yes & Yes & Yes & Yes \\
\multicolumn{1}{l}{Number of top-5\% Writers}   & Yes& Yes & Yes & Yes & Yes \\
 Adjusted $R^2$ & 0.130 & 0.131 & 0.118 & 0.139 & 0.113 \\
\bottomrule
\multicolumn{6}{p{5.1in}}{Notes: $N=645$. See table~\ref{tabl_depvars_def} for the definitions of the outcome variables. Standard Errors: (Independent: HC3) [Clustered: Candidate Univ] $|$Clustered: Univ Rank$\times$Period$|$. $^{**}$p$<$0.05; $^{***}$p$<$0.01. Candidate's characteristics: sex, region of origin, field of research, major publications, PhD lasting 7 years or longer. PhD program characteristics: rank, region.} \\
\end{tabular}

  }
\end{table}

Table~\ref{tabl_OLS_baseline_Llama_avg} shows least squares estimates of equation~\eqref{eq_OLS_avg} for the baseline outcome and with different sets of control variables.\footnote{Table~\ref{tabl_OLS_baseline_Llama_avg_controls} in the appendix presents the coefficient estimates for the control variables.} Regardless of the specifics, RL sentiment has a strong and highly statistically significant relationship with the probability of success.\footnote{Throughout the paper I present standard errors, and the corresponding levels of significance, for three different assumptions regarding the joint distribution of the error terms in equation~\eqref{eq_OLS_avg}. HC3 standard errors assume the $\epsilon_{it}$ are independently distributed but heteroskedastic. Clustered standard errors allow for dependence across errors either within PhD programs (due to norms regarding how to write RLs, for example) or within clusters defined by the market year and the PhD rank group (due to competition among candidates for similar jobs).}

Prompt-based sentiment extraction enables a straightforward interpretation of the regression results, even if they cannot be taken to measure a causal effect. For example, an increase in average polarity of 0.1 means that the LLM assigns a 10 pp. higher probability to filling the prompt with one of the positive sentiment tokens relative to the negative tokens. According to the findings in the table, such an increase is predictive of a much higher ($\sim$23 p.p.) expected probability of obtaining a top research job either in academia or at a policy institution.

In table~\ref{tabl_OLS_robust_Llama_avg}, I present similar estimates for the different definitions of success in the academic job market. The results are qualitatively identical in all cases. Among the different dependent variables, alternative~1 (removing post-docs and teaching jobs from the definition of success) seems to have a somewhat weaker dependence on RL sentiment.

\subsection{Sentiment Dispersion}

Almost without exception, employers in the job market for academic economists require interested applicants to submit \emph{multiple} confidential RLs. This convention suggests that employers are more than willing to tradeoff the extra time required to assess the additional information for insurance against the risk of biased signals from poorly incentivized letter writers.

An open question is whether disagreement among the many letter writers is reflected in worse job market outcomes for the candidates. In order to study this issue, I focus on a sub-sample of ``complete applications,'' i.e. candidates for whom I have at least three RLs including the letter from the main research adviser.\footnote{Table~\ref{tabl_OLS_comp_robust_Llama_avg} in the appendix replicates the analysis in table~\ref{tabl_OLS_robust_Llama_avg} for this sub-sample. The results are practically the same as with the full sample.}

Figure~\ref{fig_polarity_byRef} in the appendix shows that the distributions of sentiment polarity scores for different letter writers are almost identical.\footnote{In unreported analysis, I also found that the adviser's RL sentiment and other writer's RL sentiment have undistinguishable positive partial correlations with job market outcomes.} However, \emph{within-candidate} variation in RL sentiment is quite high, accounting for over 45\% of the total variance of sentiment.\footnote{An alternative way to illustrate this point is by looking at correlations across letter writer scores. These correlations range between 0.27 and 0.40. In other words, letter writers tend to be aligned with each other but are far from unanimous.}

\begin{table}
  \centering
    \caption{Sentiment Dispersion: summary stats}\label{tabl_sentiment_dispersion_stats}
    {\small
  \begin{tabular}{lcccc}
    \toprule
     & Mean & SD & Min & Max \\
    \midrule
    Sentiment Average & 0.0722 & 0.0247 & 0.0013 & 0.1579 \\
    Sentiment Range   & 0.0467 & 0.0284 & 0.0027 & 0.1505 \\
    Sentiment Mean Absolute Deviation     & 0.0172 & 0.0104 & 0.0010 & 0.0583 \\
    Sentiment Standard Deviation & 0.0233 & 0.0141 & 0.0014 & 0.0770 \\
    \bottomrule
    \multicolumn{5}{p{4.45in}}{Notes: $N$=553 (complete applications sub-sample). The range is the difference between the $\max$ and the $\min$ sentiment score for each candidate.}\\
  \end{tabular}
  }
\end{table}

Table~\ref{tabl_sentiment_dispersion_stats} shows summary statistics for alternative ways to measure sentiment dispersion among letter writers. The \emph{sentiment range}---the measure that I will use as a baseline---is defined as the difference between the $\max$ and the $\min$ sentiment score among the RLs written on behalf of a candidate. For robustness, I also consider the mean absolute deviation (MAD) and the standard deviation (SD). As is clear from the table, all measures of dispersion are quite high relative to average sentiment. They also vary significantly across candidates.

\begin{figure}[htb!]
  \centering
  \caption{Sentiment Dispersion and Success in the Academic Job Market}\label{fig_Nsuccess_rng_scatterplot}
         \includegraphics[scale=0.7]{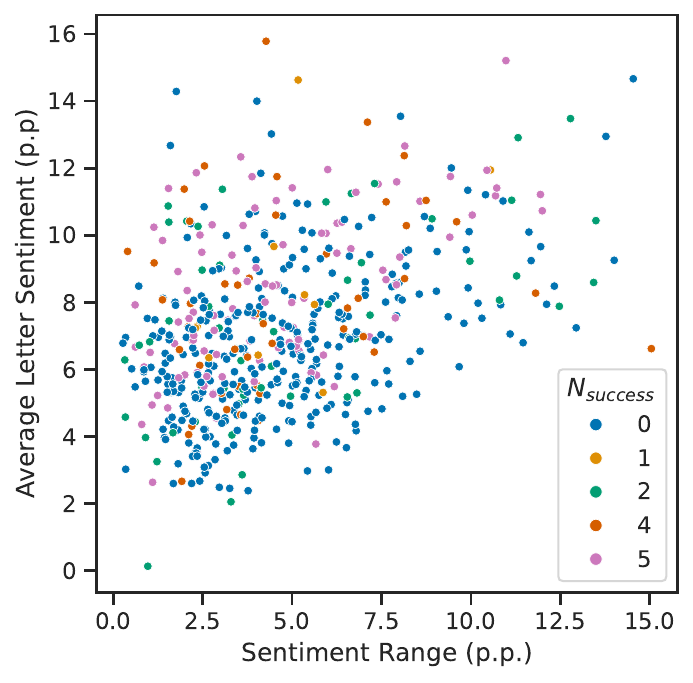}
    \vspace{-0.15in}\hspace{0.15in}
    {\small \flushleft
     \begin{tabular}{p{3.8in}}
       Notes: $N$=553 (complete application sub-sample). $N_{success}$ shows by how many alternative definitions the candidate had success in the job market. \\
     \end{tabular}
     }
\end{figure}
\vspace{0.2in}

Figure~\ref{fig_Nsuccess_rng_scatterplot} depicts the relationship between average sentiment, sentiment dispersion and success in the academic job market. Candidates who succeeded according to most definitions are clustered above and to the left relative to those who did not succeed.

To confirm this observation, I estimate the following regression:

\begin{align}\label{eq_OLS_rng}
 y_{it} =& \:\alpha+\beta_{1}\,S_{it}+\beta_{2}\,D_{it}+\boldsymbol{X}^{\prime}_{it}\boldsymbol\gamma+\epsilon_{it}
\end{align}
where $D_{it}$ is one of the measures of sentiment dispersion. Table~\ref{tabl_OLS_comp_robust_Llama_range} confirms that, regardless of the outcome definition, sentiment range has a negative relationship to success in the job market. Table~\ref{tabl_OLS_comp_baseline_Llama_dispersion} in the appendix shows similar results for alternative measures of sentiment dispersion. In summary, these estimates suggest that hiring committees fully exploit the information contained in RLs.

\begin{table}[tb!]
  \centering
  \caption{Sentiment Dispersion and Job Market Success}\label{tabl_OLS_comp_robust_Llama_range}
  {\scriptsize
  
\begin{tabular}{lccccc}
\toprule
 \multicolumn{1}{c}{\emph{Outcomes}$\rightarrow$}      & \multicolumn{1}{c}{\emph{Baseline}$^{\phantom{***}}$} & \multicolumn{1}{c}{\emph{Alt. 1}$^{\phantom{***}}$} & \multicolumn{1}{c}{\emph{Alt. 2}$^{\phantom{***}}$} & \multicolumn{1}{c}{\emph{Alt. 3}$^{\phantom{***}}$} & \multicolumn{1}{c}{\emph{Alt. 4}$^{\phantom{***}}$} \\
\midrule
 Average RL Sentiment & 0.0368$^{***}$ & 0.0245$^{***}$ & 0.0381$^{***}$ & 0.0403$^{***}$ & 0.0350$^{***}$ \\
 (p.p.)                   & (0.0092)$^{\phantom{***}}$ & (0.0079)$^{\phantom{***}}$ & (0.0094)$^{\phantom{***}}$ & (0.0107)$^{\phantom{***}}$ & (0.0089)$^{\phantom{***}}$ \\
                          & [0.0080]$^{\phantom{***}}$ & [0.0077]$^{\phantom{***}}$ & [0.0084]$^{\phantom{***}}$ & [0.0105]$^{\phantom{***}}$ & [0.0076]$^{\phantom{***}}$ \\
        & $|$0.0107$|^{\phantom{***}}$ & $|$0.0085$|^{\phantom{***}}$ & $|$0.0097$|^{\phantom{***}}$ & $|$0.0108$|^{\phantom{***}}$ & $|$0.0100$|^{\phantom{***}}$ \\[0.3cm]
 Sentiment Range ($\max-\min$) & -0.0219$^{\phantom{***}}$ & -0.0151$^{\phantom{***}}$ & -0.0229$^{\phantom{***}}$ & -0.0193$^{\phantom{***}}$ & -0.0154$^{\phantom{***}}$ \\
 (p.p.) & (0.0077)$^{***}$ & (0.0062)$^{**\phantom{*}}$ & (0.0078)$^{***}$ & (0.0084)$^{**\phantom{*}}$ & (0.0076)$^{**\phantom{*}}$ \\
 & [0.0077]$^{***}$ & [0.0060]$^{**\phantom{*}}$ & [0.0078]$^{***}$ & [0.0073]$^{***}$ & [0.0075]$^{**\phantom{*}}$ \\
 & $|$0.0073$|^{***}$ & $|$0.0051$|^{***}$ & $|$0.0074$|^{***}$ & $|$0.0077$|^{**\phantom{*}}$ & $|$0.0069$|^{**\phantom{*}}$ \\
\midrule
\multicolumn{1}{l}{Four Letters \& Avg. RL Length}& Yes& Yes & Yes & Yes & Yes \\
\multicolumn{1}{l}{Candidate's characteristics} & Yes& Yes & Yes & Yes & Yes \\
\multicolumn{1}{l}{PhD characteristics}         & Yes& Yes & Yes & Yes & Yes \\
\multicolumn{1}{l}{Job Market Year}             & Yes& Yes & Yes & Yes & Yes \\
\multicolumn{1}{l}{Number of top-5\% Writers}   & Yes& Yes & Yes & Yes & Yes \\
 Adjusted $R^2$ & 0.132 & 0.140 & 0.135 & 0.139 & 0.107 \\
\bottomrule
\multicolumn{6}{p{5.3in}}{Notes: $N=553$ (complete applications sample). See table~\ref{tabl_depvars_def} for the definitions of the outcome variables. Standard Errors: (Independent: HC3) [Clustered: Candidate Univ] $|$Clustered: Univ Rank$\times$Period$|$. $^{**}$p$<$0.05; $^{***}$p$<$0.01. Candidate's characteristics: sex, region of origin, field of research, major publications, PhD lasting 7 years or longer. PhD program characteristics: rank, region.} \\
\end{tabular}

  }
\end{table}

\subsection{Comparison of Prompt-based Sentiment with Other Approaches}

Prompt-based sentiment applied to RLs can predict success in the job market. Can other approaches reproduce this feat? To answer this question I obtained measures of sentiment polarity using each of the approaches discussed in section~\ref{section_methods}.\footnote{The HGI and LM dictionaries can be applied directly to RLs. Vader is a sentence-based sentiment analyzer, so I apply it sentence-by-sentence to every RL and then average the resulting scores. Flair and FinBert are language models with relatively small context windows (512 tokens). In order to obtain sentiment scores for the RLs, I break them into chunks and then average the scores. Table~\ref{tabl_sentiment_descriptives} in the appendix presents descriptive statistics for the different sentiment measures.}

\begin{figure}[ht!]
  \centering
    \caption{Correlation Matrix}\label{fig_correlations}
  \includegraphics[scale=0.8]{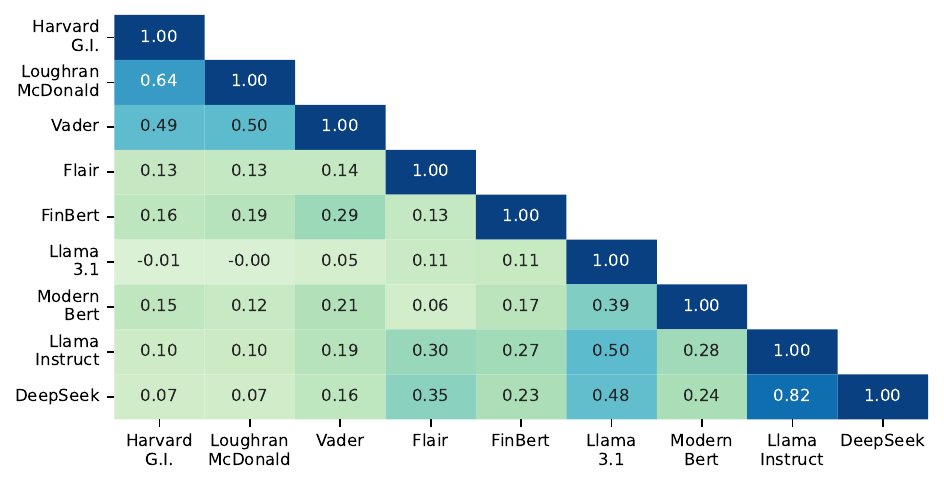}
    \vspace{-0.2in}
  {\small \flushleft \hspace{0.55in}
  \begin{tabular}{p{5.0in}}
  Notes: $N$=1968. Pearson correlation coefficients among all RL sentiment measures.\\
  \end{tabular}
  }
\end{figure}

Figure~\ref{fig_correlations} presents correlations among the different sentiment measures. While almost all correlations are positive, it is noteworthy that the coefficients are clustered by technique. For example, lexicon-based measures have much stronger associations among each other---correlations close to 0.5---than with any of the LLM-based measures. Similarly, the correlation between the two instruction-based measures is 0.82.

\begin{figure}[ht!]
  \centering
    \caption{Predicting Job Market Outcomes --- Alternative Sentiment Approaches}
    \label{fig_tstats_avg}
  \includegraphics[scale=0.7]{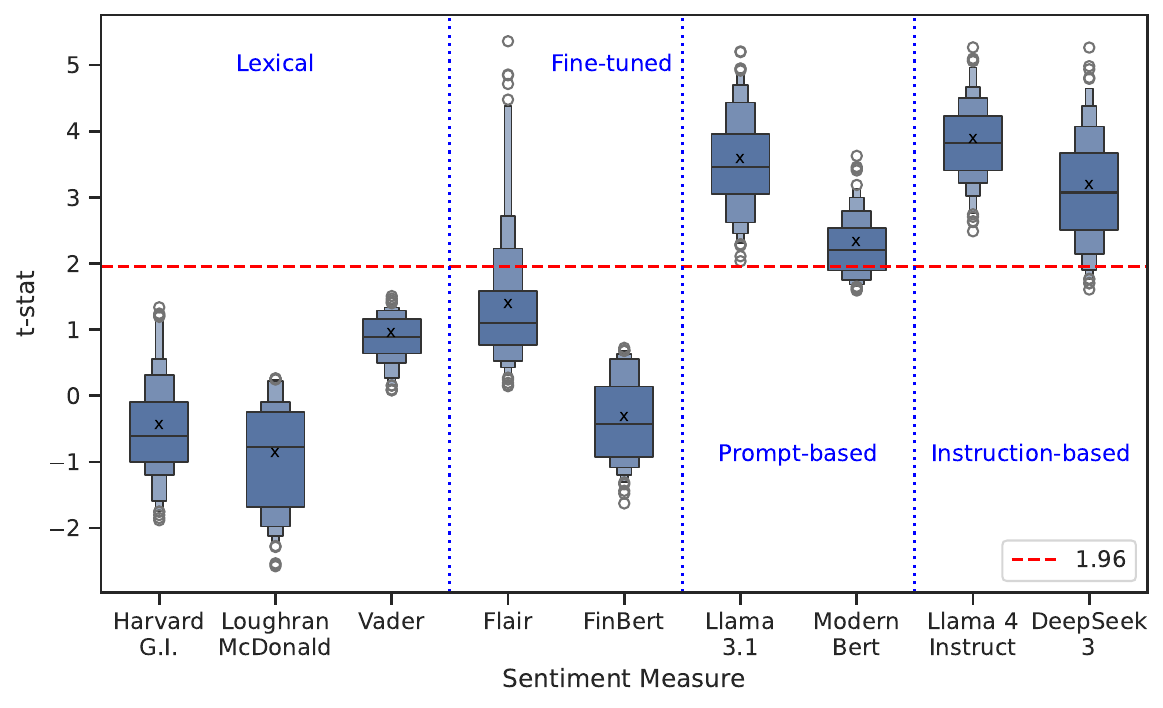}
    \vspace{-0.15in}
  {\small \flushleft \hspace{0.35in}
  \begin{tabular}{p{4.8in}}
  Notes: The box plots depict the distributions of t-statistics for the coefficient corresponding to Avg. RL Sentiment in equation~\eqref{eq_OLS_avg}. Each sentiment measure is used in 150 alternative specifications: 5 outcome variables, 2 samples, 5 control variable groups, and 3 alternative standard errors.\\
  \end{tabular}
  }
\end{figure}

I repeat the regression analysis from section~\ref{section_OLS} for each of the alternative sentiment measures. Figure~\ref{fig_tstats_avg} summarizes my findings. Specifically, the box plots in the figure show how the t-statistic for the coefficient ($\hat{\beta}$) associated with Avg.~RL~Sentiment in equation~\eqref{eq_OLS_avg} varies across specifications. For each of the sentiment measures, I estimated 150 different specifications by varying the outcome variable (5 variants), considering alternative control variable sets (5), using either the full or the restricted `complete applications sample' (2), and applying different assumptions to the error term (3).

Somewhat predictably, neither the lexical approach nor the fine-tuned LLMs are able to predict job market outcomes. The dictionaries involved in the former and the fine-tuning process in the latter target applications that are quite distant from RL sentiment extraction. In fact, in many cases the estimated coefficients obtained using these sentiment measures have the `wrong' sign.

As mentioned in section~\ref{section_methods}, ModernBert is a bidirectional LLM (encoder). These models are smaller and relatively light-weight compared to the generative LLMs. They also allow for a more flexible prompt strategy given that the [MASK] token can be included anywhere in the template. However, they suffer from a limited context window. In the case of ModernBert, the 8K token context is just enough to fit every RL in one go. However, using the full context window comes at a cost in performance. Figure~\ref{fig_alternative_scatters} (left) shows the distribution of sentiment scores when applying the prompt-based strategy using ModernBert. Comparing this figure to figure~\ref{fig_scatter_Llama31}, we can see that ModernBert produces scores that are decreasing in the length of the RL, which is the unfortunate consequence of the context size limitation.

Regardless of these limitations, the sentiment scores that result from applying ModernBert are predictive of job market outcomes. In most of the regressions, the average RL sentiment comes out positive and significant at the 5\% level.

The two instruction-based sentiment scores are also successful in predicting outcomes, with Llama~4 clearly outperforming DeepSeek~3 in this exercise. In figure~\ref{fig_alternative_scatters}, I also plot the distributions of these sentiment measures against the RL word count. In this case, the intuitive positive relationship between RL length and sentiment reappears. However, the instruction-based measures also have disadvantages. It is clear from the figure that the model generates answers that are ``round'' numbers. In addition, creating a prompt that produced sufficient variation among candidates took many repeated trials. In particular, as is clear from the instruction set I settled for, it was quite difficult to get the model to generate low scores for at least some RLs.

\begin{figure}[hbt!]
  \centering
    \caption{The Role of Sentiment Range --- Alternative Sentiment Approaches}
    \label{fig_pvalues_rng}
  \includegraphics[scale=0.7]{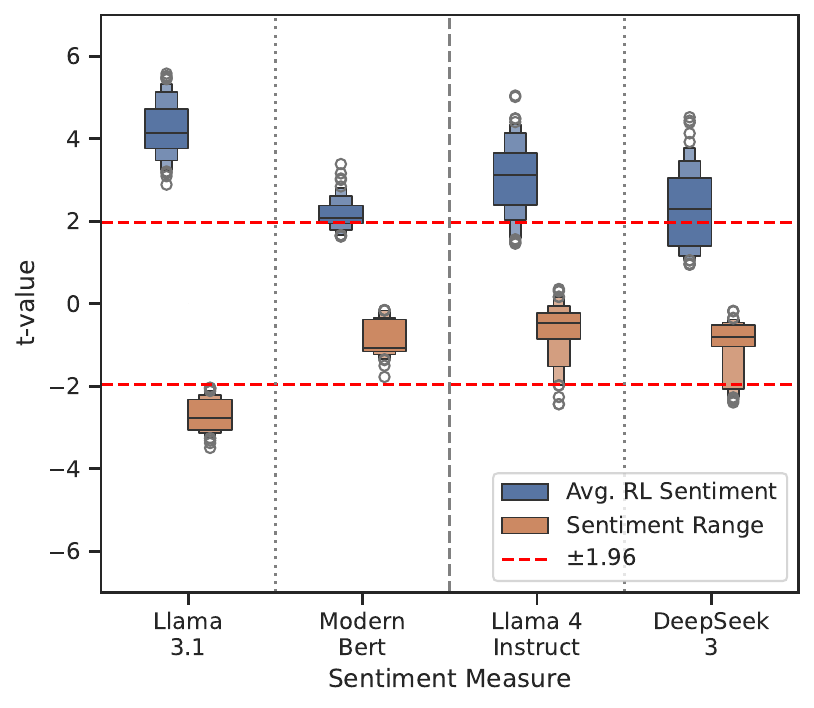}
    \vspace{-0.15in}
  {\small \flushleft \hspace{0.35in}
  \begin{tabular}{p{4.8in}}
  Notes: The box plots depict the distributions of t-statistics for the coefficients corresponding to Avg.~RL~Sentiment and RL~Sentiment~Range in equation~\eqref{eq_OLS_rng}. Each sentiment measure is used in 75 alternative specifications estimated on the `full applications' restricted sample: 5 outcome variables, 5 control variable groups, and 3 alternative standard errors.\\
  \end{tabular}
  }
\end{figure}

Figure~\ref{fig_pvalues_rng} repeats the exercise, this time for equation~\eqref{eq_OLS_rng}.\footnote{To save space, I restrict the analysis to the methods that were successful in predicting outcomes without the inclusion of sentiment dispersion.} All regressions are estimated on the `full application' restricted sample. Neither ModernBert nor the instruction-based models are able to replicate the findings in table~\ref{tabl_OLS_comp_robust_Llama_range} using the \texttt{Llama 3.1} prompt-based sentiment. Most remarkably, no other sentiment measure is able precisely to capture the role of sentiment dispersion.

\subsection{Measurement Error}

Our focus of interest is the relationship between RL sentiment and job market outcomes. However, the least squares estimates presented so far are based on a particular approach to measuring sentiment, namely my prompt-based sentiment polarity score. This approach could be subject to bias due to measurement error.

Formally, suppose that the relationship between ``true'' sentiment $S$ and the prompt-based measure $\hat{S}$ is given by $\hat{S}=S+u$, where $u$ is the measurement error. In general, simply plugging-in the mismeasured variable in equation~\eqref{eq_OLS_avg} will lead to least squares estimates that are biased (i.e. they do not consistently estimate the true population linear projection). Without making some assumptions regarding the nature of the error, however, it is not possible to say much else.\footnote{Under the classical assumptions ($\E(uS)=\E(u\epsilon)=0$), it is known that the presence of measurement error tends to \emph{attenuate} the regression coefficients.}

Recently, there have been several attempts to analyze the econometric issues that arise in applications that use variables derived from machine learning algorithms or LLMs \citep[e.g.][]{Battaglia_et_al_2024,Ludwig_et_al_2025}. The emerging consensus is that bias introduced by measurement error is indeed the main concern. The remedy suggested in this literature is to de-bias reported estimates. Unfortunately, the bias correction approach requires either a validation sample (a set of data where the variable is observed without error) or detailed knowledge regarding the process generating the error in measurement. This approach is inapplicable for an intrinsically latent variable like RL sentiment ($S$).

A feasible alternative is to take a \emph{multiple indicators} approach to addressing measurement error. This involves re-estimating the plug-in version of equation~\eqref{eq_OLS_avg} by two-stage least squares using alternative measures of sentiment as instruments for $\hat{S}$. Formally, let the instrument be given by $\tilde{S}=S+w$, where $w$ is assumed to be another classical measurement error term. The key assumption necessary for this approach to work is $\E(u w)=0$. In other words, the errors from the two sentiment measures must be uncorrelated. While the assumption is untestable, it is possible to use different instrument candidates with different measurement properties and check for robustness.

\begin{table}[tb!]
  \centering
  \caption{Two Stage Least Squares Estimates}\label{tabl_2SLS_avg}
  {\scriptsize
  \begin{tabular}{lcc}
\toprule
                          &   (1)           &   (2)               \\
Instrument ($\tilde{S}$)  &  ModernBert     &   DeepSeek3 Instruct\\
First Stage F-stat$^{\#}$  & 196.0           & 208.6               \\
\midrule
& \multicolumn{2}{l}{\textit{Baseline Outcome}} \\
Avg. RL Sentiment ($\hat{S}$) &  0.0355$^{\phantom{***}}$  &  0.0369$^{\phantom{***}}$   \\
 (p.p.)                   &  (0.0162)$^{**\phantom{*}}$ & (0.0114)$^{***}$ \\
                          &  [0.0133]$^{***}$ & [0.0111]$^{***}$ \\
                          &$|$0.0179$|^{**\phantom{*}}$ &$|$0.0132$|^{***}$ \\[0.1cm] \midrule
& \multicolumn{2}{l}{\textit{Alternative 1}} \\
Avg. RL Sentiment ($\hat{S}$) & 0.0256$^{\phantom{***}}$ & 0.0274$^{\phantom{***}}$ \\
 (p.p.)                       & (0.0137)$^{*\phantom{**}}$ & (0.0099)$^{***}$ \\
                              & [0.0120]$^{**\phantom{*}}$ & [0.0097]$^{***}$ \\
                              & $|$0.0149$|^{*\phantom{**}}$ & $|$0.0114$|^{**\phantom{*}}$ \\[0.1cm] \midrule
& \multicolumn{2}{l}{\textit{Alternative 2}} \\
Avg. RL Sentiment ($\hat{S}$) & 0.0420$^{\phantom{***}}$    & 0.0453$^{\phantom{***}}$ \\
 (p.p.)                       & (0.0167)$^{**\phantom{*}}$  & (0.0117)$^{***}$ \\
                              & [0.0140]$^{***}$ & [0.0115]$^{***}$ \\
                              & $|$0.0179$|^{**\phantom{*}}$ & $|$0.0143$|^{***}$ \\[0.1cm] \midrule
& \multicolumn{2}{l}{\textit{Alternative 3}} \\
Avg. RL Sentiment ($\hat{S}$) & 0.0373$^{\phantom{***}}$ & 0.0408$^{\phantom{***}}$ \\
 (p.p.)                       & (0.0173)$^{**\phantom{*}}$   & (0.0182)$^{**\phantom{*}}$ \\
                              & [0.0157]$^{**\phantom{*}}$   & [0.0182]$^{**\phantom{*}}$ \\
                              & $|$0.0169$|^{**\phantom{*}}$ & $|$0.0185$|^{**\phantom{*}}$ \\[0.1cm] \midrule
& \multicolumn{2}{l}{\textit{Alternative 4}} \\
Avg. RL Sentiment ($\hat{S}$) & 0.0366$^{\phantom{***}}$ & 0.0294$^{\phantom{***}}$ \\
 (p.p.)                       & (0.0156)$^{**\phantom{*}}$ & (0.0110)$^{***}$ \\
                              & [0.0132]$^{***}$ & [0.0104]$^{***}$ \\
                              & $|$0.0164$|^{**\phantom{*}}$ & $|$0.0124$|^{**\phantom{*}}$ \\
\bottomrule
\multicolumn{3}{p{3.7in}}{Notes: $N$=645. Standard Errors: (Independent: HC1) [Clustered: Candidate Univ] $|$Clustered: Univ Rank$\times$Period$|$. $^{*}$p$<$0.1; $^{**}$p$<$0.05; $^{***}$p$<$0.01. All regressions include the full set of control variables. $^{\#}$The F-stat for the significance of the excluded instrument in the first stage is computed assuming clustering at the univ.~rank~$\times$~period level.} \\
\end{tabular}

  }
\end{table}

Table~\ref{tabl_2SLS_avg} presents two-stage least squares estimates of equation~\eqref{eq_OLS_avg}. The first candidate instrument is the prompt-based sentiment polarity measure derived from applying an alternative LLM (ModernBert). On one hand, this sentiment measure applies the same prompt template and the same verbalizer.  On the other, it utilizes a bidirectional encoder-based transformer, which uses a different tokenization procedure, a different pre-training technique, and a different training set.\footnote{The architectural and methodological differences between ModernBert and Llama 3.1 are very significant. While it is reasonable to assume that much of the raw text that is the input to pre-training is shared, important details regarding the scale, variety, and the kind of pre-processing and cleaning that is applied to the raw text is unique to each project. Moreover, as mentioned above, the language modeling tasks that are the centerpiece of LLM pre-training are very different for encoder and decoders. Finally, the details of the architecture (tokenization approach and vocabulary size, type of positional embeddings, type and placement of normalization layers, type and placement of attention layers, activation functions, etc.) are extremely different.} The second instrument is the sentiment score derived from instructing DeepSeek~3. In this case, the model has been post-trained and the prompting strategy is completely different from the one used to obtain my baseline prompt-based sentiment scores. For both instruments, the F-test for the excluded instruments in the first stage regression way exceeds the critical value suggested by \citet{Stock_Yogo_2005} to rule out weak instruments.

With almost no exception, all the 2SLS estimates are positive and statistically significant.\footnote{This is somewhat remarkable considering the small sample size.} Moreover, the 2SLS estimates are sizably larger than the corresponding least squares estimates in table~\ref{tabl_OLS_robust_Llama_avg}. Specifically, the 2SLS estimates are $62\%$ larger than the OLS estimates, on average. Finally, it is noteworthy that the estimates obtained using alternative instruments are numerically quite similar. Taking these results at face value would suggest that the incidence of measurement error is substantial and that the true relationship between RL sentiment and job market outcomes is even stronger than what it might appear based on least squares regressions.

\subsection{Random Forest}

Exactly how predictive of success are the contents of RLs? In order to explore this issue, I estimate a random forest classifier with the baseline outcome as target. Besides average RL sentiment and sentiment dispersion, I include all the control variables as predictors.\footnote{See \citet{Athey_Imbens_2019} for details on random forests.} The main hyper-parameters are set based on 5-fold cross-validation.\footnote{Table~\ref{tabl_RF_hyper} in the appendix presents the hyper-parameter values. As usual, the training of each decision tree is based on a different sample bootstrap and a random subset of the predictors. I apply inverse probability weighting to address the imbalance in the outcome variable.} The random forest is trained on the `complete applications' sub-sample.

Random forests provide an easy way to assess the predictive performance of the model without the need to set aside a holdout sample. Specifically, it is possible to evaluate each tree in the forest by assessing its predictive performance on the sample elements that were not included in the bootstrap sample used to train the tree. This is referred to as `out-of-bag' (OOB) performance.

\begin{table}
  \centering
 \caption{Random Forest Out-of-Bag Performance}\label{tabl_rf_oob}
 {\scriptsize
 
\begin{tabular}{lcccc}
  \toprule
\multicolumn{1}{p{0.5in}}{Baseline Outcome}& Precision &  Recall & F1-score & Support \\
  \midrule
  Failure&   0.80    &   0.75  &   0.77   &   402   \\
  Success&   0.42    &   0.48  &   0.45   &   151   \\
  \midrule
  Accuracy     &      &         &   0.68   &   553   \\
  Macro Avg.   & 0.61 &   0.62  &   0.61   &   553   \\
  Weighted Avg.& 0.69 &   0.68  &   0.69   &   553   \\
  \bottomrule
\end{tabular}

 }
\end{table}

Table~\ref{tabl_rf_oob} presents the OOB performance metrics for the prediction of the baseline outcome. Clearly, predicting job market outcomes is much harder than predicting text meta-data. Regardless, the overall accuracy score of 0.68 is not horrible for a random forest trained on 553 examples.

\begin{figure}[htb!]
  \centering
  \caption{Random Forest Permutation Importances}\label{fig_RF_permutation_importance}
  \includegraphics[scale=0.75]{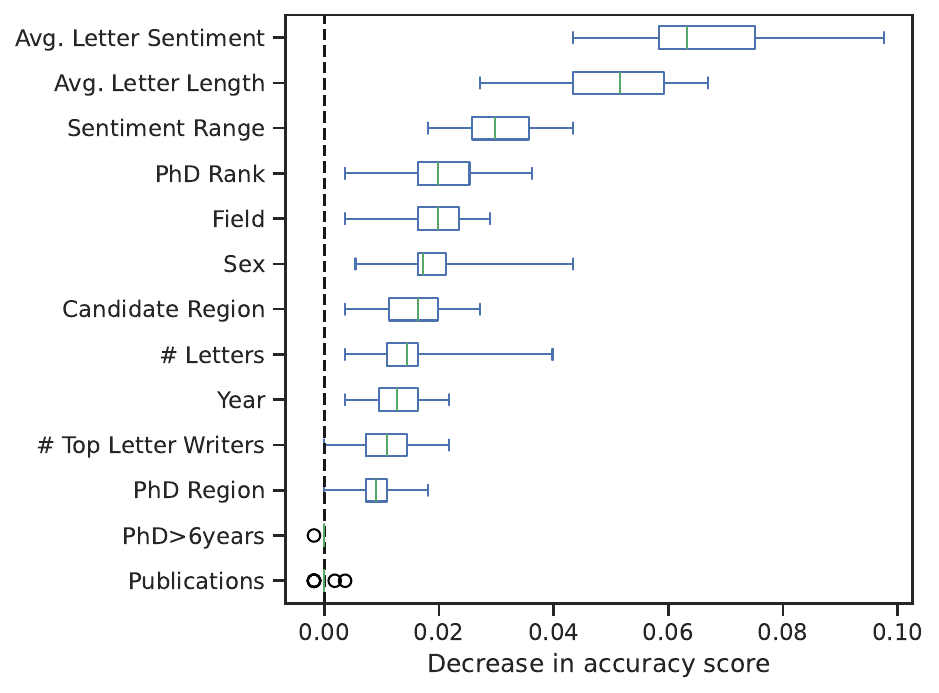}
  \vspace{-0.15in}
    {\small \flushleft \hspace{1.375in}
    \begin{tabular}{p{3.1in}}
     Notes: The box plots depict the accuracy decrease distributions for 30 independent re-shuffles of each predictive feature.\\
    \end{tabular}
    }
\end{figure}

Figure~\ref{fig_RF_permutation_importance} presents the relative importance of different predictive features. Specifically, I use permutation importances, which measure the average decrease in the OOB accuracy score when the corresponding feature is shuffled.\footnote{Specifically, the feature is re-shuffled and the random forest re-trained 30 times.} The average RL sentiment score is the most important predictive feature, with a median decrease in accuracy above 6~p.p. Average RL length and the sentiment range come second and third, respectively. Remarkably, other predictor variables usually thought to be very important (the PhD rank, publications, the top-ranked letter writers, having a `good job market year') are not highly predictive of success.

The random forest is also a way to relax the linearity assumption. Specifically, I use the trained forest to obtain partial dependence curves, which simulate the changes in the predicted probability of the outcome for counterfactual values of each feature. Formally, the partial dependence curve for the predictor $X_S$ is obtained as follows:
\begin{align*}
  \mathrm{pd}_{X_S}(x_S)=\mathrm{E}_{X_C}\left[\hat{P}(x_S,X_C) \right], \quad x_S \in \mathcal{X}_S
\end{align*}
where $X_C$ are all other predictors and $\mathcal{X}_S$ is the support set of $X_S$. It is important to emphasize that the $\mathrm{pd}$ curves try to approximate the conditional expectation function without assuming linearity. They do \emph{not} provide causal estimates.

\begin{figure}[hb!]
  \centering
  \caption{Random Forest Partial Dependence Curves}
  \label{fig_pd_RL}
        \centering
  	    \includegraphics[scale=0.75]{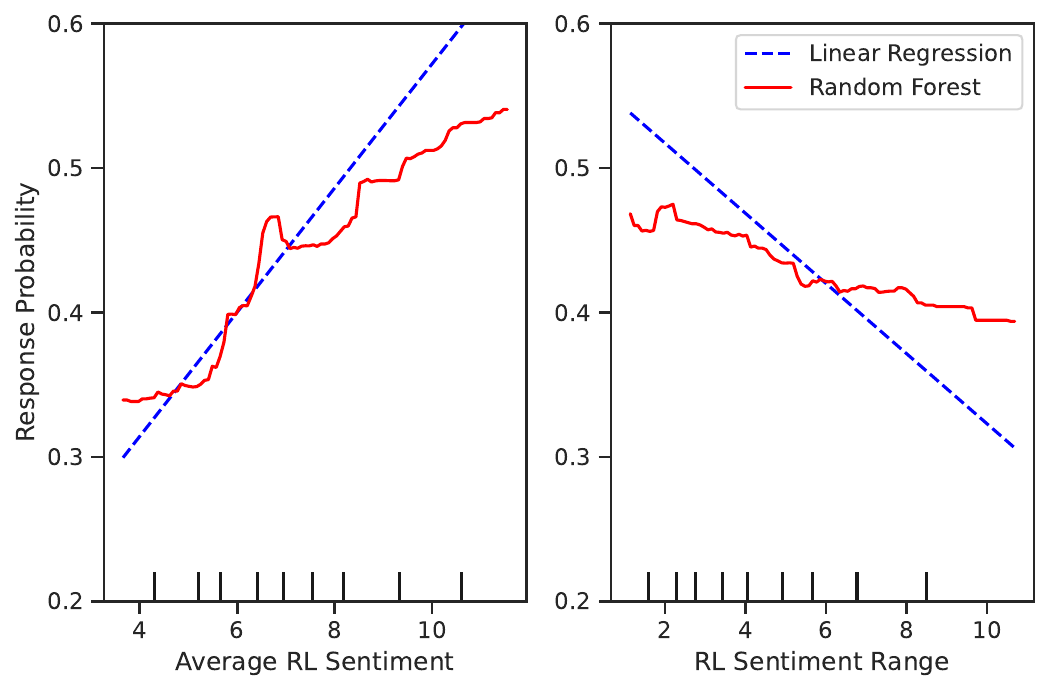}
    \vspace{-0.3in}
    {\small \flushleft \hspace{0.4in}
    \begin{tabular}{p{5.85in}}
    Notes: Partial dependence curves simulate average job market outcomes when the whole sample is assigned counterfactual levels of the predictor. The vertical marks on the \emph{x}-axis show deciles of the predictor's distribution.\\
    \end{tabular}
    }
\end{figure}

In figure~\ref{fig_pd_RL}, I present $\mathrm{pd}$ curves for average RL sentiment and sentiment range. For comparison, I also plot the corresponding simulation based on a linear regression.\footnote{The regression in the plot is not exactly the same as the regression in the first column of table~\ref{tabl_OLS_comp_robust_Llama_range} because I use inverse probability weights in order to match the random forest fit.} While the random forest confirms the positive association between average sentiment scores and employment outcomes, it suggests that the relationship might not be linear. In particular, the curve seems to have steep jumps after crossing certain thresholds. The random forest also points to the relationship between RL sentiment range and job market success being less pronounced than the regression coefficients suggest.

\begin{figure}[hbt!]
  \centering
  \caption{Partial Dependence Contours}
  \label{fig_pd_interactions_wcXsentXrng}
    \begin{subfigure}{0.49\textwidth}
    \centering
    \includegraphics[scale=0.565]{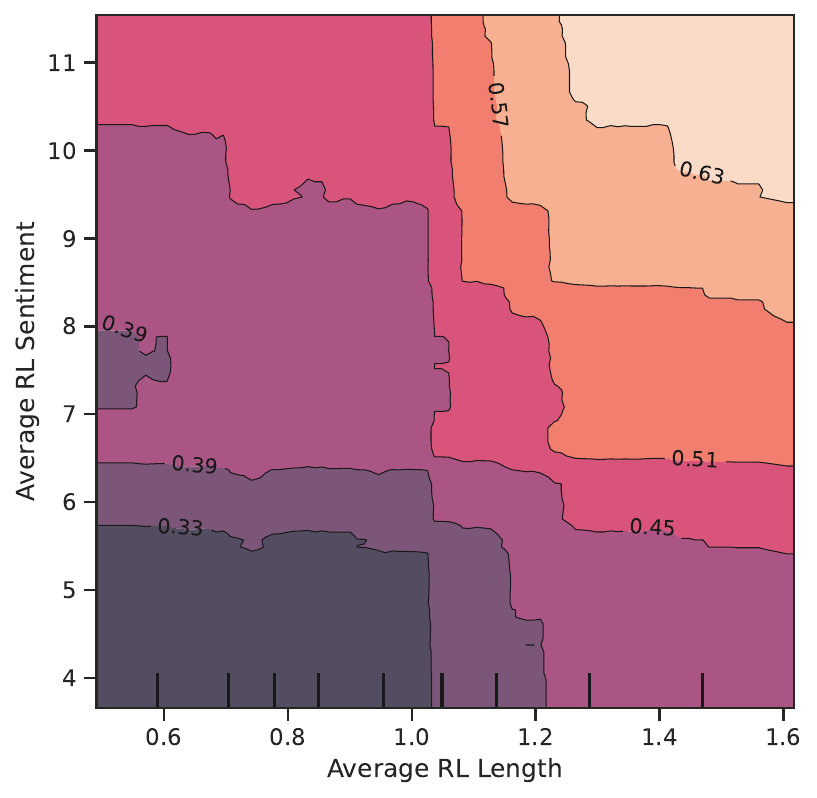}
    \end{subfigure}
    \begin{subfigure}{0.49\textwidth}
    \centering	\includegraphics[scale=0.565]{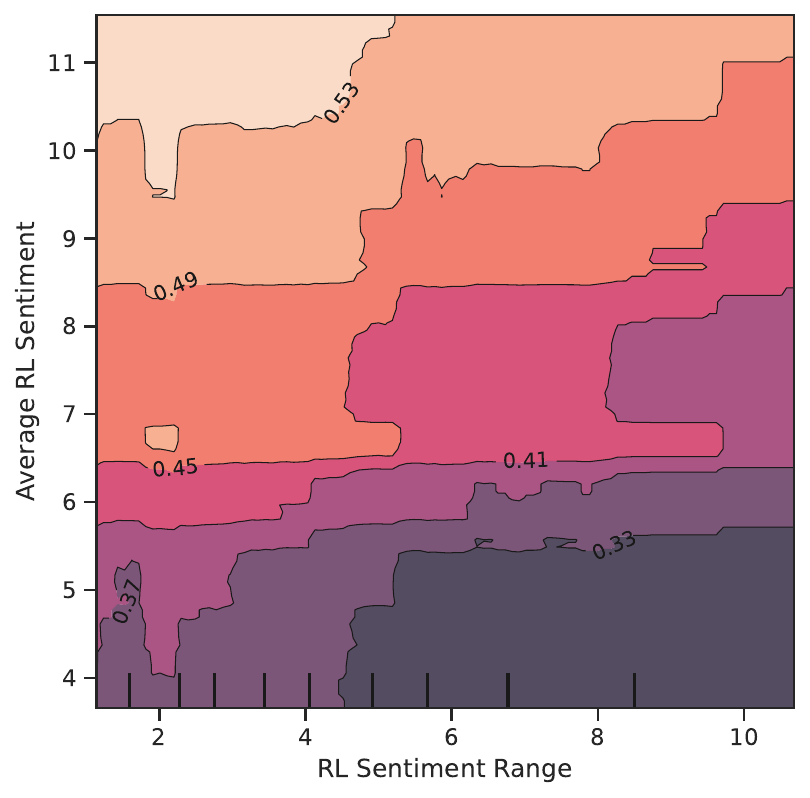}
    \end{subfigure}
    \vspace{-0.15in}
    {\small \flushleft \hspace{0.2in}
    \begin{tabular}{p{5.8in}}
    Notes: Partial dependence contours simulate average job market outcomes when the whole sample is assigned  counterfactual levels of two continuous predictors simultaneously. The level curves depict different probabilities of success in the academic job market.\\
    \end{tabular}
    }
\end{figure}

The same method can be applied to study interactions between predictors. Figure~\ref{fig_pd_interactions_wcXsentXrng} (left) shows the average probabilities of success when both RL length and sentiment are assigned counterfactual levels. Clearly, these two signals positively interact, with the lowest probability of success in the lower left corner and the highest probability of success in the upper right corner.

The plot on the right of the figure shows simulations for Avg. RL sentiment and RL sentiment range. In this case, the highest probability of success is predicted for candidates with high average RL sentiment and very low sentiment range. Conversely, the lower right corner of the plot has the lowest predicted probability of success.

\begin{figure}[ht!]
  \centering
  \caption{Partial Dependence Interactions}
  \label{fig_pd_interactions_sexfieldetc}
  \begin{subfigure}{0.49\textwidth}
  \centering
  \caption{Sex}  \includegraphics[scale=0.65]{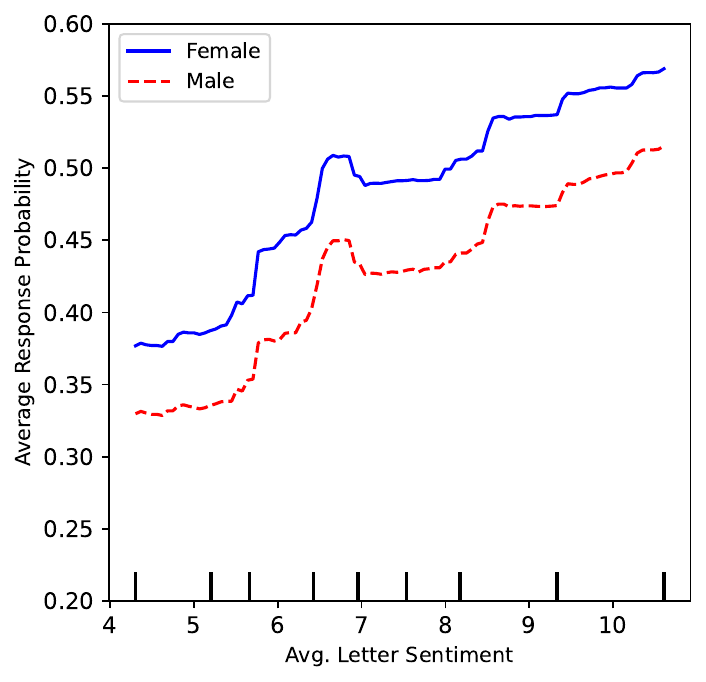}
  \end{subfigure}
  \begin{subfigure}{0.49\textwidth}
  \centering
  \caption{Candidate Region}  \includegraphics[scale=0.65]{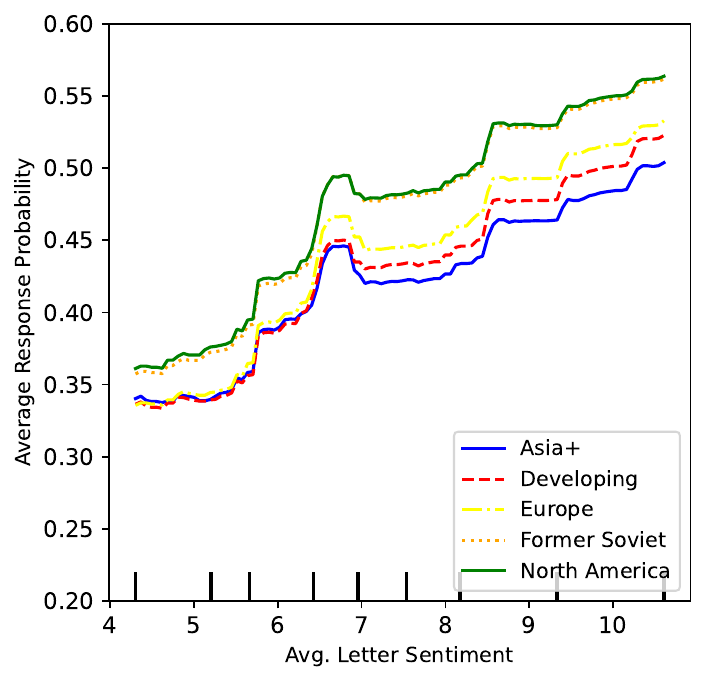}
  \end{subfigure}
  \begin{subfigure}{0.49\textwidth}
  \centering
  \caption{Field of Research}  \includegraphics[scale=0.65]{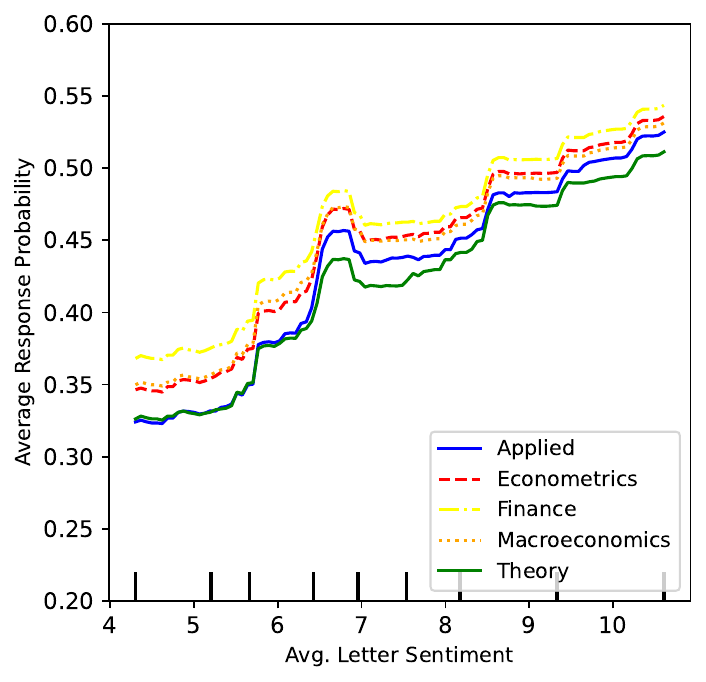}
  \end{subfigure}
  \begin{subfigure}{0.49\textwidth}
  \centering
  \caption{PhD Rank}  \includegraphics[scale=0.65]{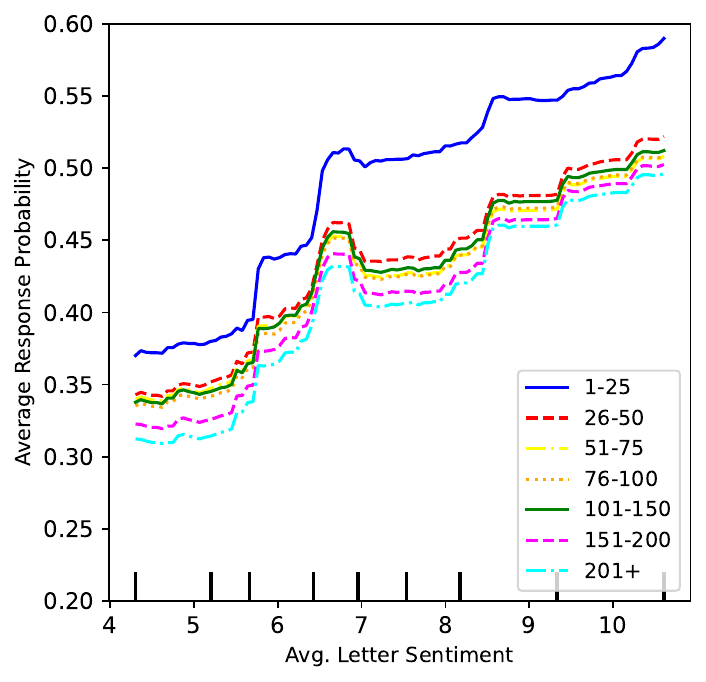}
  \end{subfigure}
  \vspace{-0.15in}
  {\small \flushleft \hspace{0.2in}
  \begin{tabular}{p{5.7in}}
  Note: Partial dependence interactions simulate average job market outcomes when the whole sample is assigned counterfactual values for Avg. RL sentiment and one discrete predictor.\\
  \end{tabular}
  }
\end{figure}

The plots in figure~\ref{fig_pd_interactions_sexfieldetc} present similar simulations, this time by assigning counterfactual levels to Avg. RL sentiment and four different discrete predictors:

\begin{itemize}
  \item[(a)] Female candidates in my sample generally perform better than males. Everyone benefits from positive RLs at more or less the same rate. In section~\ref{section_gender} below, I show that the real picture is less rosy when considering other aspects of RL content.
  \item[(b)] There is a clear pattern that seems to strongly disfavor candidates from Asia. Not only do they have a lower probability of success, but also seem to benefit less from positive RLs.
  \item[(c)] There seems to be a small difference in rates of success across fields of research for low average sentiment levels. Finance candidates perform a bit better than applied people and theorists. However, these differences become smaller as RL sentiment increases.
  \item[(d)] In contrast, the random forest predictions suggest that the ranking of the PhD program makes a very sizeable difference in terms of obtaining a research position. Specifically, candidates from the top-25 programs have markedly higher success rates, and the gap is strongly increasing with RL sentiment. Candidates from these programs that manage to secure RLs in the top decile of sentiment have an almost 60\% predicted probability of attaining a position in a top research institution.
\end{itemize}

\section{Gendered Sentiment}\label{section_gender}

Prompt-based learning is a general technique for text analysis. In this section, I show how it can be deployed to extract information regarding the candidate's main strengths, as evaluated by the letter writer.

There is a large literature specialized on seemingly harmless differences in how women and men are described. Within economics, \citet{Eberhardt_et_al_2023}, \citet{Baltrunaite_et_al_2024}, and \citet{Hochleitner_et_al_2025} have shown that the language used to describe women in RLs accentuates what they call `grindstone' aspects of their personality. These are traits such as conscientiousness, patience, attention to detail, being a good communicator and collaborator, etc. Male candidates, on the other hand, are more often described with `standout' traits such as originality, leadership, technical excellence, etc. Theses small differences in language are not really harmless in their effects in the job market.

The papers in this literature mostly follow a lexical approach.\footnote{The approach in \citet{Baltrunaite_et_al_2024} is somewhat different in that they use embeddings.} They have curated dictionaries of terms associated with personality traits, and then classified RLs based on the relative frequency with which terms in the different dictionary categories are used in the RL text. In this section, I show how a prompt-based strategy can be used to extract analogous information.

\begin{table}[hbt!]
  \centering
    \caption{Prompt-based Extraction of Personality Traits}
    \label{tabl_prompt_grindstone}
    {\small
    \begin{tabular}{m{1.4in}p{0.4in}m{3.8in}}
      \toprule
      \multicolumn{1}{c}{\textbf{Prompt} ($\mathcal{T}_{personality}$)} & \multicolumn{2}{c}{\textbf{Verbalizer} ($\mathcal{V}_{personality}$)} \\
      \midrule
      ``[X] If I had to pick this candidate's main strength, I would mention their [MASK]'' &&
          \begin{itemize}
          \item[$\textbf{standout}$:] \small{\{ability, research, creativity, originality, leadership, skill, skills, intelligence, theoretical, technical, mathematical, analytical, statistical, intellectual, original, critical, ambition, drive, excellence\}}
          \item[$\textbf{grindstone}$:] \small{\{passion, dedication, diligence, patience, perseverance, persistence, open, personality, genuine, interpersonal, collaborative, enthusiasm, work, attention, humility, communication, maturity, flexibility, commitment, writing, teaching\}}
        \end{itemize}
     \\
      \bottomrule
    \end{tabular}
    }
\end{table}

Table~\ref{tabl_prompt_grindstone} presents the  template/verbalizer combination that I employed. A key distinctive feature of my approach is that the model is implicitly forced to choose which aspect of the candidate's personality to emphasize. For the sake of brevity, throughout this section I focus mostly on the difference between the probabilities assigned to the two types of traits:
$$\mathrm{P}\left(y=\text{standout} \mid \boldsymbol x\right) - \mathrm{P}\left(y=\text{grindstone}\mid \boldsymbol x \right).$$%
For lack of a better term, I refer to this quantity as the \emph{net-standout score}.

Figure~\ref{fig_net-standout_scatters} in the appendix shows the distribution of the net-standout score. The distribution is symmetrical and has a remarkably wide range, going from strongly negative (grindstone-biased) to highly positive (strongly standout-biased) RLs.\footnote{Summary statistics for the net-standout score can be found in table~\ref{tabl_sentiment_descriptives} in the appendix.} Similarly to the prompt-based sentiment polarity measure, net-standout is clearly increasing in RL length. Remarkably, however, the scatter plot of sentiment polarity and net-standout has no distinguishable pattern. The correlation coefficient between these two scores is $0.13$. In summary, net-standout measures a completely independent feature of the information contained in RLs.

\begin{figure}[hbt!]
  \centering
    \caption{Net-Standout Score --- Gender Differences}
    \label{fig_net-standout-genderdiff}
  \includegraphics[scale=0.6]{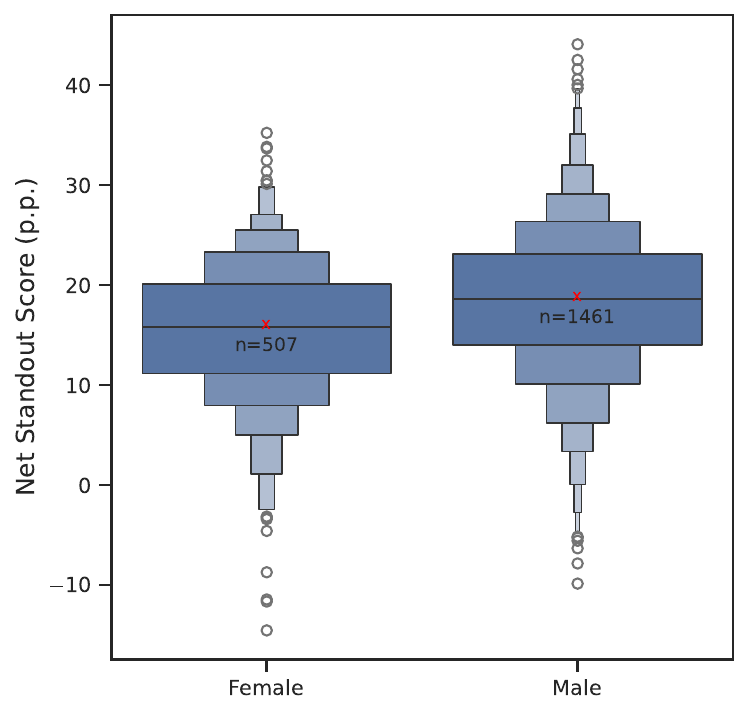}
    \vspace{-0.15in}
  {\small \flushleft \hspace{0.35in}
  \begin{tabular}{p{3in}}
  Notes: $N$=1968. The box plots show the distribution of the net-standout score (in p.p.) for female and male candidates. The difference in means is $2.82$ and the HC3 standard error is $0.376$.\\
  \end{tabular}
  }
\end{figure}

Are there differences in net-standout scores between male and female candidates? As is clear from figure~\ref{fig_net-standout-genderdiff}, the answer is a rotund `yes'. Similar to the literature, I find that RLs written for male candidates emphasize `standout' traits more, whereas females are more likely to be described using `grindstone' attributes. Moreover, the gender effect is robust to introducing controls for other candidates characteristics, for the PhD rank and region, for the job market year, and for a host of letter writer characteristics including gender, whether the writer is the main adviser, whether the writer is either a top-10 or a top-5 percentile author, and the writer's title and potential experience.\footnote{These regression results are available in the online appendix (TODO).}

\begin{figure}[hbt!]
  \centering
  \caption{Predicting Net-Standout --- Partial Dependence Tables}
  \label{fig_pd_net-standout_tables}
  \begin{subfigure}{\textwidth}
  \centering
  \includegraphics[scale=0.75]{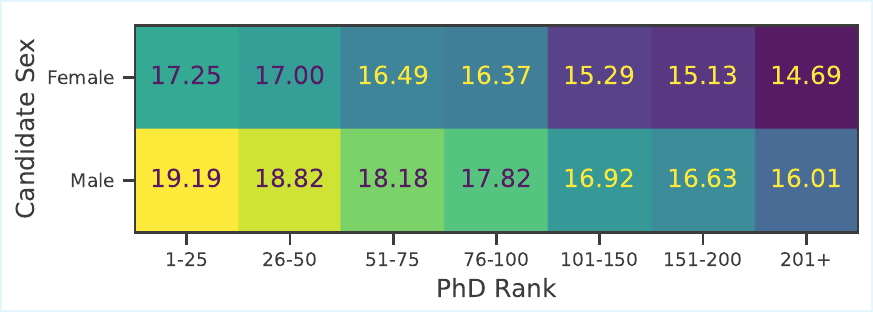}
  \end{subfigure}\\
  \begin{subfigure}{0.39\textwidth}
  \centering
  \includegraphics[scale=0.675]{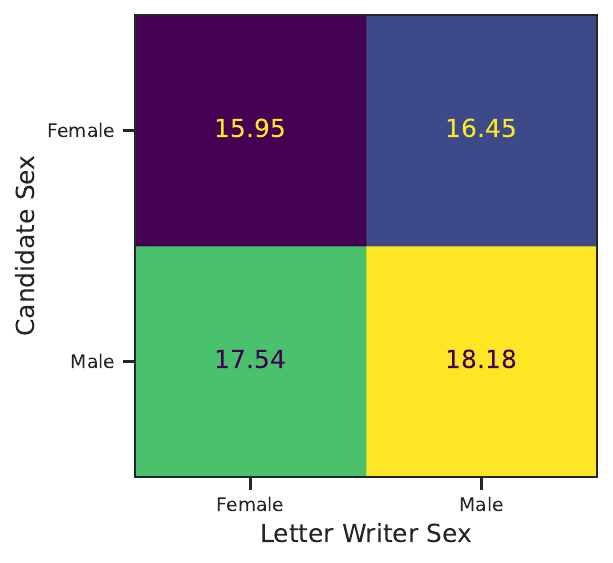}
  \end{subfigure}
  \begin{subfigure}{0.59\textwidth}
  \centering
  \includegraphics[scale=0.7]{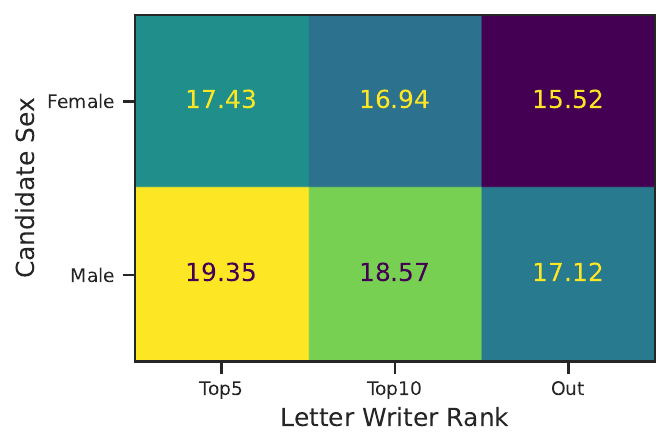}
  \end{subfigure}
  \vspace{-0.15in}
  {\small \flushleft \hspace{0.2in}
  \begin{tabular}{p{5.6in}}
  Note: Partial dependence tables simulate the RL net-standout score when the whole sample is assigned counterfactual values for the candidate's sex and one additional discrete predictor.\\
  \end{tabular}
  }
\end{figure}

\begin{figure}[htb!]
    \centering
    \caption{Predicting Net-Standout --- Letter Writer Experience}
    \label{fig_pd_net-standout_ref_exper}
    \includegraphics[scale=0.7]{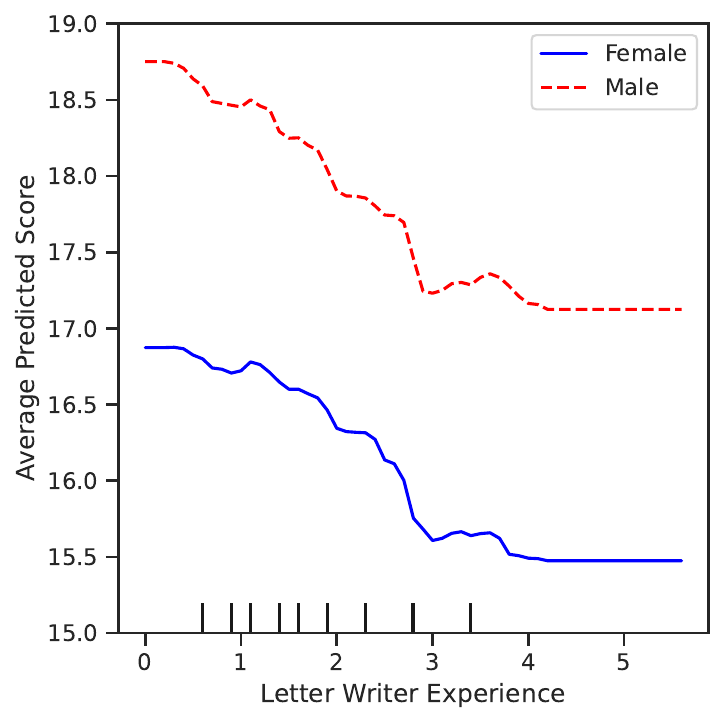}
  \vspace{-0.15in}
  {\small \flushleft \hspace{0.2in}
  \begin{tabular}{p{4in}}
  Note: Partial dependence interaction simulates the RL net-standout score when the whole sample is assigned counterfactual values for the candidate's sex and the letter writer's potential experience (in decades).\\
  \end{tabular}
  }
\end{figure}

Using a random forest, I once again explore interactions, this time between the candidate's gender and other potential predictors of the net-standout score.\footnote{Full details on this procedure, including hyper-parameters and performance metrics are available in the online appendix (TODO).} The top panel of figure~\ref{fig_pd_net-standout_tables} shows the interaction between candidate gender and the rank of their PhD program. The gender gap in net-standout is present throughout the ranking distribution. The table implies that a male candidate from a PhD program ranked 76--100 is likely to get a higher net-standout score than a female in a top-25 program.

The table in the bottom right panel reinforces this point. Regardless of the prestige of the letter writer, women are more likely to be described with grindstone terms. Finally, the bottom left panel shows that female candidates cannot expect to receive a higher net-standout score if they pick a female referee.

Figure~\ref{fig_pd_net-standout_ref_exper} is meant to provide some hint as to whether we can expect this gender gap to disappear in the near future. It shows the interaction of candidate sex and letter writer potential experience (in decades) in explaining the net-standout score. While younger scholars seem to be more inclined to emphasize candidates' ability over other traits, there is no evidence that the gap is any narrower for letter writers who received their PhDs recently.

In appendix table~\ref{tabl_OLS_robust_gendered}, I present regression results when equation~\eqref{eq_OLS_avg} is augmented to include `gendered' sentiment variables. Specifically, I include the average standout ($\mathrm{P}\left(y=\text{standout} \mid \boldsymbol x\right)$) and average grindstone ($\mathrm{P}\left(y=\text{grindstone}\mid \boldsymbol x \right)$) scores as predictors of job market outcomes. The coefficients for average RL sentiment are almost identical to those in table~\ref{tabl_OLS_robust_Llama_avg}. The coefficients for the average standout score are relatively small and statistically insignificant. The average grindstone score, on the other hand, has a negative and statistically significant association with the probability of success.\footnote{The exception is again the outcome that excludes post-docs from the definition of success.}

\section{Conclusion}

The lexical approach to sentiment analysis has limited power. Because dictionaries cannot capture context, the resulting sentiment measures are unable to extract some relevant information. The prompt-based strategy described in this paper is a simple alternative that leverages the enormous resources invested in pre-training LLMs. Prompt-based sentiment extraction offers numerous advantages:

\begin{enumerate}
  \item The text input can be entered ``as is'', without the need of any heavy-handed pre-processing. Multiple languages are acceptable.
  \item LLMs take context into account.
  \item No labeled data and no fine-tuning is necessary.
  \item The resulting sentiment scores have a probability interpretation.
  \item The model performance can be benchmarked against known meta-data.
  \item The technique can be extended to capture text features other than general sentiment.
\end{enumerate}

In addition to this ``static'' advantages, it should be kept in mind that LLMs are becoming more powerful at a very fast pace. As a result, the advantages of prompt-based learning over the alternatives can only be expected to grow over time.

However, this approach to sentiment analysis is not without its challenges. Prompt and verbalizer engineering can be time-consuming. For certain applications, coming up with an appropriate prompt may be challenging. And implementing prompt-based learning is certainly not a trivial programming task.

Some researchers will be more inclined to simply query an instruction-based model \emph{\`a la} Chat-GPT. I show that the output from such an exercise has to always be taken with a pinch of salt. These models pass the full output vector to a text-generating function that effectively samples a single token from the whole vocabulary.

In order to showcase the technique, in this paper I apply prompt-based sentiment extraction to a small administrative dataset of confidential RLs. As a form of validation, I show that relatively simple prompts can predict known candidate meta-data such as sex and field of research.

Using the resulting sentiment scores, I show that there is compelling evidence that RLs are predictive of job market outcomes. The candidate's average letter polarity is strongly predictive of success in the job market, regardless of the precise definition of the outcome variable. Moreover, I show that disagreement among the letter writers is penalized by the market.

All in all, the evidence seems to be roughly consistent with the canonical \citet{Saloner_1985} model. While there is competition among referees to place their students in good institutions, the repetitive nature of the interaction between letter-writers and employers means that there are counterbalancing forces. The resulting equilibrium is one in which relatively stronger candidates get better recommendations, employers place significant trust in the contents of the RLs, and those candidates are more likely to get the best job offers. This meritocratic mechanism seems to operate quite broadly, across fields of research, personal characteristics of the candidate and throughout the rankings (both of PhD programs and of letter writers).

On a less positive note, using a different prompt I am able to show that there is a persistent gap in how male and female candidates are described in RLs. Specifically, RLs written for females tend to be somewhat biased towards `grindstone' personality traits. This gender gap, which exists regardless of field, PhD program rank and letter-writer characteristics, is likely one of the contributing factors to the continuing under-representation of women in the profession.

\newpage
\bibliography{references2}

\newpage\clearpage
\appendix\label{dataappendix}
\section{Appendix}
\setcounter{table}{0}
\renewcommand\thetable{\Alph{section}.\arabic{table}}
\setcounter{figure}{0}
\renewcommand\thefigure{\Alph{section}.\arabic{figure}}

\begin{table}[htb!]
  \centering
  \caption{Descriptive Statistics --- Alternative Sentiment Measures}
  \label{tabl_sentiment_descriptives}
  {\scriptsize
  \begin{tabular}{lccc|cc}
\toprule
&\multicolumn{3}{c|}{\emph{Lexical Approach}}   & \multicolumn{2}{c}{\emph{Fine-tuned LLMs}}\\
 & Harvard & Loughran &\multicolumn{1}{c|}{Vader}& Flair & FinBert  \\
 & G.I.    & McDonald & \multicolumn{1}{c|}{}    &       &          \\
\midrule
Mean      & 0.516  & 0.088& 0.257&0.982 &  0.396   \\
Std. Dev. & 0.157  & 0.304& 0.098&0.098 &  0.445   \\
Min       & -0.089 &-1.000&-0.153&0.004 & -0.826   \\
Median    & 0.520  & 0.080& 0.259&1.000 &  0.094   \\
Max       & 1.000  & 1.000& 0.740&1.000 &  1.000   \\[0.15cm]
\toprule
&\multicolumn{3}{c|}{\emph{Prompt-based Sentiment}}   &  \multicolumn{2}{c}{\emph{Instruction-tuned LLMs}}        \\
     & Llama& Modern&Net-Standout& Llama~Instruct & DeepSeek  \\
     & 3.1  & Bert  &Score& 4/16E          &  3        \\
\midrule
Mean      &0.073 & 0.282& 0.176& 0.696 & 0.822 \\
Std. Dev. &0.034 & 0.167& 0.076& 0.219 & 0.094 \\
Min       &-0.008& 0.000&-0.145& 0.000 & 0.200 \\
Median    &0.068 & 0.256& 0.179& 0.800 & 0.800 \\
Max       &0.221 & 0.868& 0.441& 1.000 & 0.900 \\
\bottomrule
\multicolumn{6}{p{4.3in}}{Notes: $N$=1968. See Section~\ref{section_methods} for an explanation of each of these approaches to sentiment extraction. The net-standout score is introduced in section~\ref{section_gender}.}\\
\end{tabular}

  }
\end{table}

\begin{table}[ht!]
  \centering
  \caption{Predicting Success in the Academic Job Market: control variables}\label{tabl_OLS_baseline_Llama_avg_controls}
  {\scriptsize
  
\begin{tabular}{lc|lc}
\toprule
\multicolumn{4}{c}{\emph{Baseline Outcome}: top-100 academic and top-50 policy jobs}  \\
\multicolumn{4}{c}{(model 5)}  \\
\midrule
 \textbf{Sex}: Male &  -0.1155$^{***}$ & \textbf{PhD Region}: USA & -0.0623 \\
                    &  (0.0415) [0.0401] $|$0.0444$|$        &               &  (0.0388) [0.0320]$^{*}$ $|$0.0361$|^{*}$   \\
\midrule
\multicolumn{2}{l|}{\textbf{Candidate's Region} (ref. Asia)} & \multicolumn{2}{l}{\textbf{Field} (ref. Applied)}\\
 Developing Countries   &  0.0953$^{}$                       & Econometrics      &  0.1411   \\
                        & (0.0657) [0.0623] $|$0.0825$|$     &   &  (0.0812)$^{*}$ [0.0716]$^{**}$ $|$0.0851$|^{*}$ \\
 Europe                 &  0.0507$^{}$                       & Finance           &  0.1694$^{***}$ \\
                        &  (0.0507) [0.0377] $|$0.0377$|$    &   &  (0.0470) [0.0450] $|$0.0448$|$       \\
 Former USSR            &  0.0653$^{}$                       & Macroeconomics   &  0.1050  \\
                        &  (0.0505) [0.0466] $|$0.0424$|$    &   &  (0.0473)$^{**}$ [0.0452]$^{**}$ $|$0.0402$|^{***}$ \\
 North America          &  0.2204$^{**}$                     & Theory             &  -0.0725$^{}$   \\
                        &  (0.1125) [0.1035] $|$0.0952$|$    &   &  (0.0553) [0.0560] $|$0.0585$|$       \\
\midrule
 \textbf{Publication(s)}  & 0.0964$^{}$ & \textbf{PhD Length}  &  0.0192$^{}$\\
                          & (0.0782) [0.0720] $|$0.0648$|$    & $\le$6 years   &  (0.0589) [0.0559] $|$0.0436$|$   \\
\midrule
\multicolumn{2}{l|}{\textbf{PhD University Rank} (ref. 1--25)}&\multicolumn{2}{l}{\textbf{Top-5\% Letter Writers} (ref. 0)}\\
  26--50   &  -0.1345                              &  One         & -0.0017$^{}$ \\
           &  (0.0543)$^{**}$ [0.0393]$^{***}$ $|$0.0564$|^{**}$ &    &  (0.0431) [0.0426] $|$0.0392$|$\\
  51--75   &  -0.1268                              &  Two or more &  0.0291$^{}$ \\
           &  (0.0696)$^{*}$ [0.0632]$^{**}$ $|$0.0568$|^{**}$    &    &  (0.0501) [0.0436] $|$0.0536$|$  \\  \cline{3-4}
  76--100  &  -0.1286$^{*}$   &  \multicolumn{2}{l}{\textbf{Job Market Year} (omitted)$^{\#}$}  \\\cline{3-4}
           &  (0.0706) [0.0710] $|$0.0665$|$         &  \multicolumn{2}{l}{\textbf{Number of Letters} (ref. One)}  \\
  101--150 &  -0.0856$^{}$                  &   Two    & -0.2439  \\
           &  (0.0640) [0.0576] $|$0.0642$|$&          & (0.1282)$^{*}$ [0.1299]$^{*}$ $|$0.0799$|^{***}$    \\
  151--200 &  -0.1368$^{*}$                &   Three  & -0.1877  \\
           &  (0.0716) [0.0568] $|$0.0649$|$&          & (0.1231)$^{}$ [0.1270]$^{}$ $|$0.0919$|^{**}$   \\
  201+     &  -0.1473                       &    Four  & -0.0817$^{}$ \\
           &  (0.0671)$^{**}$ [0.0567]$^{***}$ $|$0.0721$|^{**}$& & (0.1272) [0.1314] $|$0.0912$|$     \\ \midrule
\textbf{Avg. RL Length}      &  0.2447$^{***}$  && \\
 (000s words)       &  (0.0537) [0.0546] $|$0.0515$|$ && \\
 \bottomrule
 \multicolumn{4}{p{5.5in}}{Notes: This table contains estimates for the control variables corresponding to model (5) in table~\ref{tabl_OLS_baseline_Llama_avg}. Standard Errors: (Independent: HC3) [Clustered: Candidate University] $|$Clustered: Univ Rank$\times$Period$|$. $^{*}$p$<$0.1; $^{**}$p$<$0.05; $^{***}$p$<$0.01. $^{\#}$None of the job market year coefficients were statistically significant.} \\
\end{tabular}

  }
\end{table}

\begin{figure}[ht!]
  \centering
  \caption{Prompt-based Sentiment Distribution for Different Letter Writers}\label{fig_polarity_byRef}
  \includegraphics[scale=0.7]{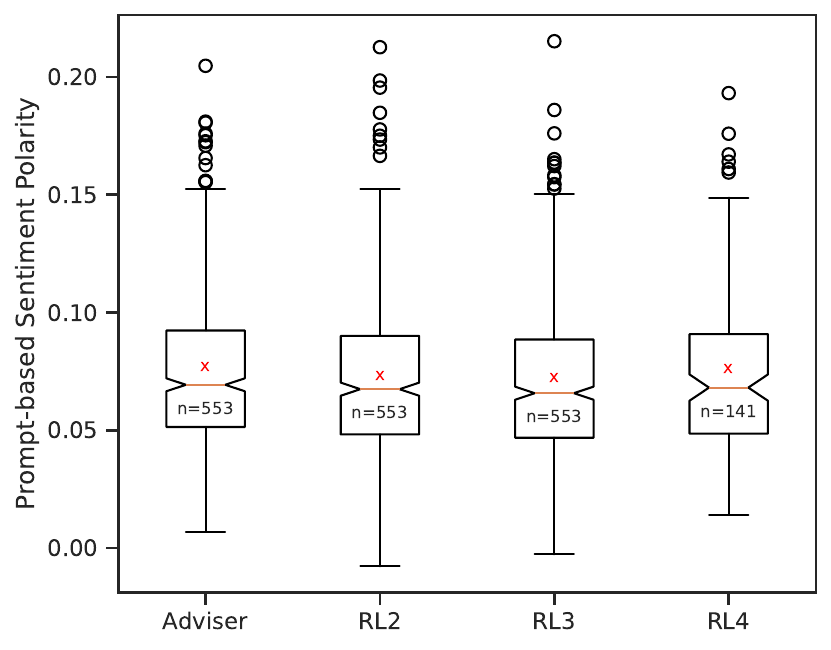}
    \vspace{-0.15in}
    {\small \flushleft 
     \begin{tabular}{p{3.4in}}
       Notes: $N$=553 (complete applications sub-sample). The order of letter writers other than the main adviser has been randomized. \\
     \end{tabular}
     }
\end{figure}

\begin{table}[tb!]
  \centering
  \caption{Predicting Success: complete applications sub-sample}\label{tabl_OLS_comp_robust_Llama_avg}
  {\scriptsize
  
\begin{tabular}{lccccc}
\toprule
 \multicolumn{1}{c}{\emph{Outcomes}$\rightarrow$}      & \multicolumn{1}{c}{\emph{Baseline}$^{\phantom{***}}$} & \multicolumn{1}{c}{\emph{Alt. 1}$^{\phantom{***}}$} & \multicolumn{1}{c}{\emph{Alt. 2}$^{\phantom{***}}$} & \multicolumn{1}{c}{\emph{Alt. 3}$^{\phantom{***}}$} & \multicolumn{1}{c}{\emph{Alt. 4}$^{\phantom{***}}$} \\
\midrule
 Average Letter Sentiment & 0.0266$^{\phantom{***}}$ & 0.0174$^{\phantom{***}}$ & 0.0274$^{\phantom{***}}$ & 0.0313$^{\phantom{***}}$ & 0.0279$^{\phantom{***}}$ \\
 (p.p.)                   & (0.0082)$^{***}$ & (0.0071)$^{**\phantom{*}}$ & (0.0084)$^{***}$ & (0.0095)$^{***}$ & (0.0079)$^{***}$ \\
        & [0.0068]$^{***}$ & [0.0064]$^{***}$ & [0.0073]$^{***}$ & [0.0092]$^{***}$ & [0.0066]$^{***}$ \\
        & $|$0.0105$|^{**\phantom{*}}$  & $|$0.0083$|^{**\phantom{*}}$  & $|$0.0091$|^{***}$  & $|$0.0105$|^{***}$ & $|$0.0096$|^{***}$ \\
\midrule
\multicolumn{1}{l}{RL Count \& Avg. RL Length}  & Yes& Yes & Yes & Yes & Yes \\
\multicolumn{1}{l}{Candidate's characteristics} & Yes& Yes & Yes & Yes & Yes \\
\multicolumn{1}{l}{PhD characteristics}         & Yes& Yes & Yes & Yes & Yes \\
\multicolumn{1}{l}{Job Market Year}             & Yes& Yes & Yes & Yes & Yes \\
\multicolumn{1}{l}{Number of top-5\% Writers}   & Yes& Yes & Yes & Yes & Yes \\
 Adjusted $R^2$ & 0.119 & 0.132 & 0.121 & 0.131 & 0.101 \\
\bottomrule
\multicolumn{6}{p{5.1in}}{Notes: $N=553$ (complete applications sample). See table~\ref{tabl_depvars_def} for the definitions of the outcome variables. Standard Errors: (Independent: HC3) [Clustered: Candidate Univ] $|$Clustered: Univ Rank$\times$Period$|$. $^{**}$p$<$0.05; $^{***}$p$<$0.01. Candidate's characteristics: sex, region of origin, field of research, major publications, PhD lasting 7 years or longer. PhD program characteristics: rank, region.} \\
\end{tabular}

  }
\end{table}

\begin{table}[tb!]
  \centering
  \caption{Sentiment Dispersion and Job Market Success: alternative measures}\label{tabl_OLS_comp_baseline_Llama_dispersion}
  {\scriptsize
  \begin{tabular}{lccc}
\toprule
 &  \multicolumn{3}{c}{\emph{Baseline~Outcome}}\\
 & \multicolumn{1}{c}{(1)} & \multicolumn{1}{c}{(2)} & \multicolumn{1}{c}{(3)}  \\
\midrule
 Average Letter Sentiment & 0.0368$^{***}$ & 0.0366$^{***}$ & 0.0360$^{***}$ \\
 (p.p.) & (0.0087) & (0.0093) & (0.0092) \\
        & [0.0080] & [0.0080] & [0.0079] \\
        & $|$0.0107$|$ & $|$0.0108$|$ & $|$0.0108$|$ \\[0.3cm]
 Sentiment Range (max-min) & -0.0219$^{***}$ & & \\
 (p.p.) & (0.0077) & & \\
        & [0.0077] & & \\
        & $|$0.0073$|$ & & \\[0.3cm]
 Sentiment Standard Deviation & & -0.0419$^{***}$ & \\
 (p.p.) & & (0.0153) & \\
        & & [0.0151] & \\
        & & $|$0.0147$|$ & \\[0.3cm]
 Sentiment Mean Absolute Deviation & & & -0.0532$^{***}$ \\
 (p.p.) & & & (0.0204) \\
        & & & [0.0201] \\
        & & & $|$0.0200$|$ \\
\midrule
\multicolumn{1}{l}{Four Letters \& Avg. RL Length}& Yes& Yes & Yes \\
\multicolumn{1}{l}{Candidate's characteristics} & Yes& Yes & Yes \\
\multicolumn{1}{l}{PhD characteristics}         & Yes& Yes & Yes \\
\multicolumn{1}{l}{Job Market Year}             & Yes& Yes & Yes \\
\multicolumn{1}{l}{Number of top-5\% Writers}   & Yes& Yes & Yes \\
 Adjusted $R^2$ & 0.132 & 0.131 & 0.130 \\
\bottomrule
\multicolumn{4}{p{4in}}{Notes: $N=553$ (complete applications sub-sample). Standard Errors: (Independent: HC3) [Clustered: Candidate Univ] $|$Clustered: Univ Rank$\times$Period$|$. $^{***}$p$<$0.01. Candidate's characteristics: sex, region of origin, field of research, major publications, PhD lasting 7 years or longer. PhD program characteristics: rank, region.} \\
\end{tabular}

  }
\end{table}

\begin{table}[ht!]
  \centering
  \caption{Random Forest Hyper-Parameters}\label{tabl_RF_hyper}
  {\scriptsize
  \begin{tabular}{lc}
\toprule
  \texttt{Max Depth} & 6 \\
  \texttt{Min Elements to Split Node} & 21 \\
  \texttt{Min Elements in Leaf Node} & 8 \\
  \texttt{Forest Size} & 120 \\ \bottomrule
  \multicolumn{2}{p{3.5in}}{Notes: Hyper-parameters result from 5-fold CV search. Samples are \texttt{bootstrapped} and $\sqrt{13}\approx 4$ predictors are selected at random in the CART tree estimation. The impurity measure is the Gini. Observations are weighted by the inverse probability of the outcome to address class imbalance.}
\end{tabular}

  }
\end{table}

\begin{figure}[hbt!]
  \centering
  \caption{The Distribution of Alternative Sentiment Measures}
  \label{fig_alternative_scatters}
  \includegraphics[scale=0.675]{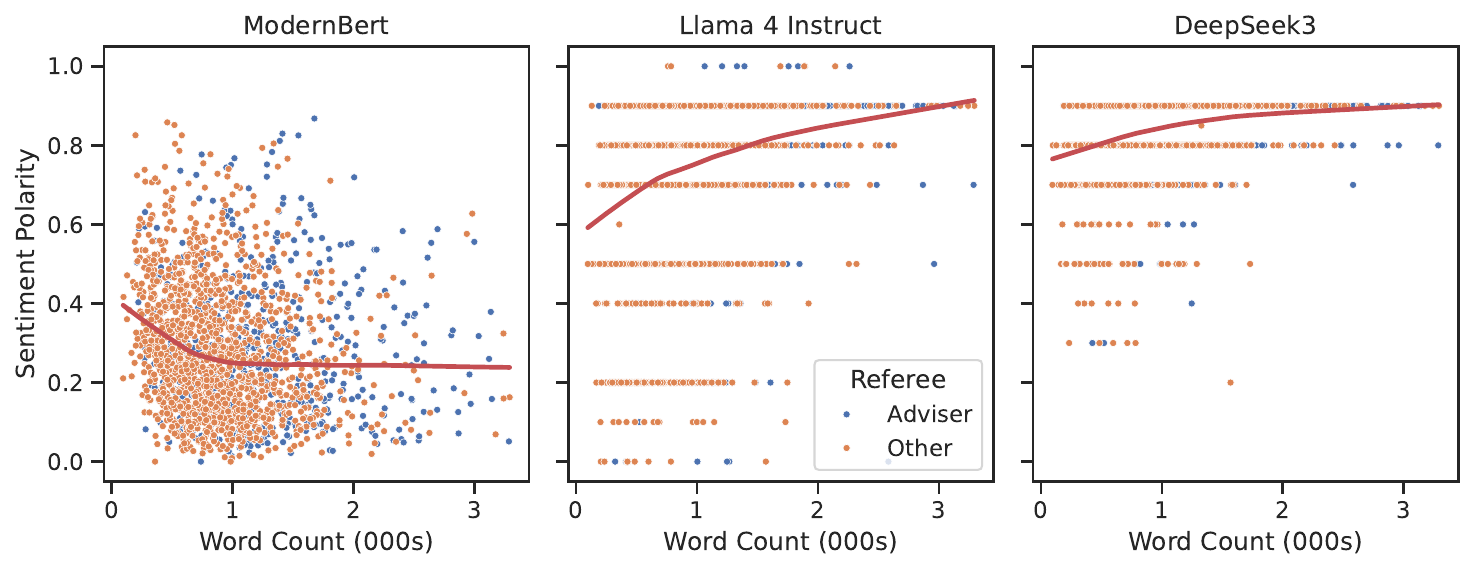}
  \vspace{-0.35in}
  {\small \flushleft \hspace{-0.05in}
  \begin{tabular}{p{5.2in}}
  Note: $N$=1968. Fitted line is a locally weighted regression.\\
  \end{tabular}
  }
\end{figure}

\begin{figure}[hbt!]
  \centering
  \caption{The Distribution of the Net-Standout Score}
  \label{fig_net-standout_scatters}
  \includegraphics[scale=0.6]{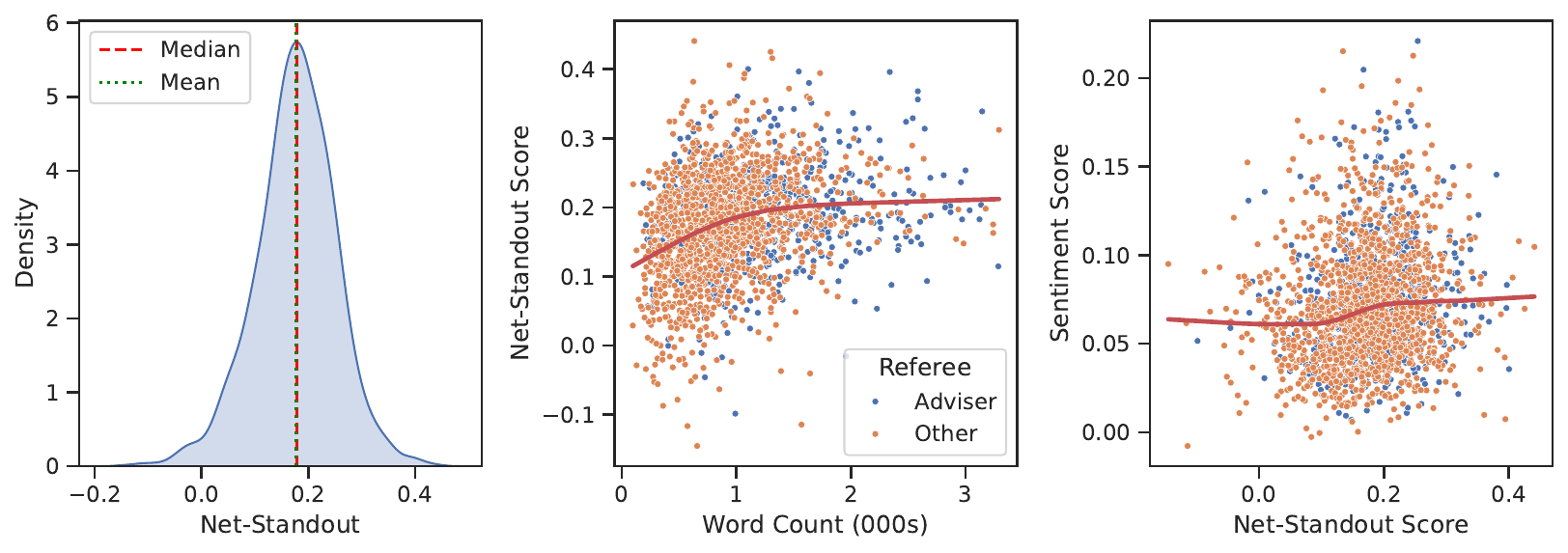}
  \vspace{-0.35in}
  {\small \flushleft \hspace{-0.05in}
  \begin{tabular}{p{5.2in}}
  Note: $N$=1968. Fitted line is a locally weighted regression.\\
  \end{tabular}
  }
\end{figure}

\begin{table}[tb!]
  \centering
  \caption{Gendered Sentiment and Job Market Outcomes}\label{tabl_OLS_robust_gendered}
  {\scriptsize
  
\begin{tabular}{lccccc}
\toprule
 \multicolumn{1}{c}{\emph{Outcomes}$\rightarrow$}      & \multicolumn{1}{c}{\emph{Baseline}$^{\phantom{***}}$} & \multicolumn{1}{c}{\emph{Alt. 1}$^{\phantom{***}}$} & \multicolumn{1}{c}{\emph{Alt. 2}$^{\phantom{***}}$} & \multicolumn{1}{c}{\emph{Alt. 3}$^{\phantom{***}}$} & \multicolumn{1}{c}{\emph{Alt. 4}$^{\phantom{***}}$} \\
\midrule
 Average RL Sentiment & 0.0215$^{***}$ & 0.0135$^{**\phantom{*}}$ & 0.0220$^{***}$ & 0.0258$^{***}$ & 0.0220$^{***}$ \\
 (p.p.)               & (0.0075)$^{\phantom{**}}$ & (0.0065)$^{\phantom{*}}$ & (0.0078)$^{\phantom{**}}$ & (0.0082)$^{\phantom{**}}$ & (0.0070)$^{\phantom{**}}$ \\
                      & [0.0065]$^{\phantom{**}}$ & [0.0057]$^{\phantom{*}}$ & [0.0069]$^{\phantom{**}}$ & [0.0081]$^{\phantom{**}}$ & [0.0064]$^{\phantom{**}}$ \\
                      & $|$0.0102$|^{**}$ & $|$0.0075$|^{*}$ & $|$0.0090$|^{**}$ & $|$0.0102$|^{**}$ & $|$0.0098$|^{**}$ \\[0.3cm]
 Average Standout Score & -0.0054$^{\phantom{***}}$ & 0.0058$^{\phantom{***}}$ & -0.0070$^{\phantom{***}}$ & -0.0103$^{\phantom{***}}$ & -0.0101$^{\phantom{***}}$ \\
 (p.p.) & (0.0076) & (0.0067) & (0.0078) & (0.0082) & (0.0070)$^{\phantom{*}}$ \\
        & [0.0062] & [0.0061] & [0.0066] & [0.0067] & [0.0064]$^{\phantom{*}}$ \\
        & $|$0.0057$|^{}$ & $|$0.0065$|^{}$ & $|$0.0066$|^{}$ & $|$0.0067$|^{}$ & $|$0.0060$|^{*}$ \\[0.3cm]
 Average Grindstone Score & -0.0179$^{\phantom{***}}$ & -0.0043$^{\phantom{***}}$ & -0.0246$^{\phantom{***}}$ & -0.0289$^{\phantom{***}}$ & -0.0213$^{\phantom{***}}$ \\
 (p.p.) & (0.0098)$^{*\phantom{*}}$ & (0.0084)$^{}$ & (0.0101)$^{**\phantom{*}}$ & (0.0106)$^{***}$ & (0.0096)$^{**\phantom{*}}$ \\
        & [0.0080]$^{**}$ & [0.0068]$^{}$ & [0.0086]$^{***}$ & [0.0091]$^{***}$ & [0.0078]$^{***}$ \\
        & $|$0.0077$|^{**}$ & $|$0.0074$|^{}$ & $|$0.0078$|^{***}$ & $|$0.0090$|^{***}$ & $|$0.0081$|^{***}$ \\
\midrule
\multicolumn{1}{l}{Number of Letters \& Avg. RL Length}  & Yes& Yes & Yes & Yes & Yes \\
\multicolumn{1}{l}{Candidate's characteristics} & Yes& Yes & Yes & Yes & Yes \\
\multicolumn{1}{l}{PhD characteristics}         & Yes& Yes & Yes & Yes & Yes \\
\multicolumn{1}{l}{Job Market Year}             & Yes& Yes & Yes & Yes & Yes \\
\multicolumn{1}{l}{Number of top-5\% Writers}   & Yes& Yes & Yes & Yes & Yes \\
 Adjusted $R^2$ & 0.132 & 0.132 & 0.124 & 0.147 & 0.117 \\
\bottomrule
\multicolumn{6}{p{5.6in}}{Notes: $N=645$. See table~\ref{tabl_depvars_def} for the definitions of the outcome variables. Standard Errors: (Independent: HC3) [Clustered: Candidate Univ] $|$Clustered: Univ Rank$\times$Period$|$. $^{*}$p$<$0.1; $^{**}$p$<$0.05; $^{***}$p$<$0.01. Candidate's characteristics: sex, region of origin, field of research, major publications, PhD lasting 7 years or longer. PhD program characteristics: rank, region.} \\
\end{tabular}

  }
\end{table}

\end{document}